\documentclass[10pt,twocolumn,letterpaper]{article}

\usepackage{iccv}
\usepackage{times}
\usepackage{epsfig}
\usepackage{graphicx}
\usepackage{amsmath}
\usepackage{amssymb}
\usepackage{booktabs}
\usepackage{multirow}
\usepackage{multicol}
\usepackage{makecell}

\usepackage[pagebackref=true,breaklinks=true,letterpaper=true,colorlinks,bookmarks=false]{hyperref}

\newcommand{\dataset}{HQ-50K}
\iccvfinalcopy 

\def\iccvPaperID{} 

\ificcvfinal\fi

\begin{document}

\title{\dataset: A Large-scale, High-quality Dataset for Image Restoration}

\author{Qinhong Yang$^{1}$
\quad Dongdong Chen$^{2}$ \quad Zhentao Tan$^1$ \quad Qiankun Liu$^1$ 
\quad Qi Chu$^{1}$ \\ \quad Jianmin Bao$^2$ 
\quad Lu Yuan$^2$ 
\quad Gang Hua$^{3}$,
\quad Nenghai Yu$^{1}$\\
{$^3$University of Science and Technology of China}  \quad  {$^2$Microsoft} \quad {$^3$Xi'an Jiaotong University} \\
\small{\texttt{\{qhyang233@mail., tzt@mail., liuqk3@mail., qchu@, ynh@\}ustc.edu.cn}}\\
\small{\texttt{\{jianbao, luyuan\}@microsoft.com}},\quad \small{\texttt{\{cddlyf, ganghua\}@gmail.com}}
}

\maketitle
\ificcvfinal\thispagestyle{empty}\fi

\begin{abstract}
This paper introduces a new large-scale image restoration dataset, called \textbf{\dataset}, which contains 50,000 high-quality images with rich texture details and semantic diversity. We analyze existing image restoration datasets from five different perspectives, including data scale, resolution, compression rates, texture details, and semantic coverage. However, we find that all of these datasets are deficient in some aspects. In contrast, \dataset~considers all of these five aspects during the data curation process and meets all requirements. We also present a new Degradation-Aware Mixture of Expert (DAMoE) model, which enables a single model to handle multiple corruption types and unknown levels. Our extensive experiments demonstrate that \dataset~ consistently improves the performance on various image restoration tasks, such as super-resolution, denoising, dejpeg, and deraining. Furthermore, our proposed DAMoE, trained on our \dataset, outperforms existing state-of-the-art unified models designed for multiple restoration tasks and levels. The dataset and code are available at \url{https://github.com/littleYaang/HQ-50K}.
\end{abstract}

\section{Introduction}
\label{sec:intro}
Large-scale high-quality datasets play an essential role in the deep learning era, which act as the catalyst stimulating and accelerating of technique development. In the image understanding field, there are many large-scale datasets proposed in the past years, \eg., ImageNet \cite{ImageNet} and COCO \cite{lin2014microsoft}, making great contribution to the field. And the data scale is still continuing growing up. But for the low-level image restoration field, even though some datasets have been collected for specific restoration tasks, \eg., super-resolution (SR)~\cite{DIV2K,Flickr2K}, denoising~\cite{BSD400,BSD432} and deraining~\cite{rain100,rain800}, there is still no large-scale dataset dedicatedly developed. By analyzing existing restoration datasets from five aspects, \ie., data scale, image resolution, compression rates, texture details and semantic coverage, we find that all existing restoration datasets are deficient in some aspects. 

In this paper, we move a step forward and propose a large-scale, high-quality dataset \dataset~for image restoration, which considers the above five aspects simultaneously. 

\begin{figure}[!t]
  \centering
  \includegraphics[width=0.45\textwidth]{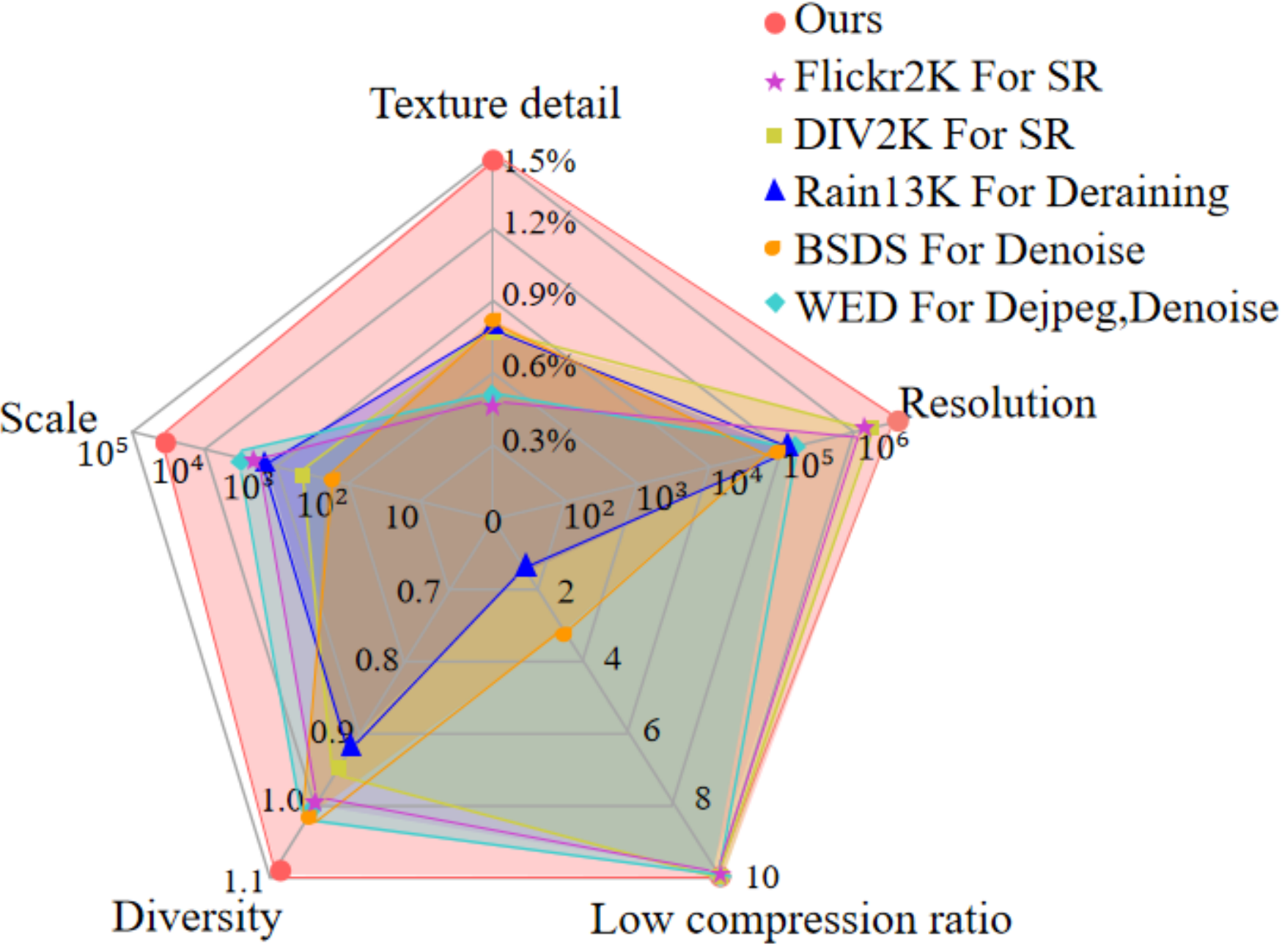}
  \caption{Comparison between our \dataset~ and existing restoration datasets from five aspects. Our dataset considers all five aspects simultaneously while existing datasets are always deficient in some aspects.}
  \label{fig:motivation}
\end{figure}

\begin{itemize}
    \item \textit{Large-Scale.} Our \dataset~contains 50,000 images, while the commonly used restoration dataset BSDS400 \cite{BSD400}, DIV2K~\cite{DIV2K}, Flickr2K~\cite{Flickr2K}, WED~\cite{WED}, and Rain13K~\cite{mprnet} only contains 400, 800, 2650, 4744 and 1212 clean images respectively. Such a larger scale can help better exploit the full potentials of more advanced model architectures.
    \item \textit{High-Resolution.} \dataset~contains high-resolution images with an average of 2,509,509 pixels, making them superior in resolution compared to other existing datasets. By utilizing \dataset~, restoration models can enhance their ability to learn more realistic textures, particularly for tasks that involve predicting high-resolution images, such as SR.
    \item \textit{Compression Rates.} Many images on the internet use compression techniques like JPEG compression to save storage space. To ensure that the images we select have minimal compression distortion, we only choose those with low compression rates by measuring the bpp metric (\ie.,``bits per pixel"). Similar to resolution, this selection criteria is important to improve the performance of restoration tasks such as SR and Dejpeg.
    \item \textit{Rich texture details.} In addition to compression, images on the internet also have an imbalanced frequency distribution, \ie., a large portion of images lack rich texture details (\eg., simple background). So indiscriminately scaling up data by crawling high-resolution images without selection will lead to an unbalanced dataset. Training the model on such data will result in poor performance when dealing with images that have intricate texture details. To overcome this, we perform frequency analysis and ensure that we select images from different frequency bands to achieve a balanced distribution when building \dataset~. 
    \item \textit{Semantic Coverage.} Existing restoration datasets, such as DIV2K \cite{DIV2K} and Flick2K \cite{Flickr2K}, primarily comprise outdoor natural images. We notice that models trained on these datasets do not perform as well for indoor or artificial images, such as maps, posters, and handwriting. To develop a more generalized restoration model, we collect images that cover a broader range of semantic categories in a more balanced way.
\end{itemize}

Following the common synthesis for learning pipeline, \dataset can be used to serve different restoration tasks by simulating the corresponding degradation process upon the clean images. Additionally, we also offer 1250 test images that span across various semantic categories and frequency ranges. This new benchmark can facilitate detailed and fine-grained performance comparison and analysis.

In addition to \dataset~, we propose a new model called the Degradation-Aware Mixture of Expert (DAMoE), which enables a unified approach to handle various restoration tasks and unknown degradation levels. To our knowledge, this is the first attempt at integrating the Mixture of Expert (MoE) concept into the image restoration field. The DAMoE model shares all the modules and parameters for different tasks, except for the experts in the MoE layers. It dynamically adjusts the inference process of the network through expert mechanism to adapt to different restoration task. For the MoE layers, we introduce two types of MoE blocks, namely, Hard-MoE and Soft-MoE, and demonstrate that their combination yields the best overall performance.
Our contributions can be summarized as two-folds:
\begin{itemize}
    \item We propose a large-scale, high-quality dataset for image restoration tasks. Through extensive experiments, we demonstrate that the proposed dataset can improve the restoration performance for different tasks.
    \item We present the first attempt at integrating MoE into the image restoration tasks and propose a new unified image restoration model ``DAMoE". It achieves better performance than existing unified restoration models. 
\end{itemize}

\section{Related Works}
\noindent\textbf{Image Restoration Datasets.}
For low-level image processing tasks, it is often very difficult and expensive to collect paired images from the real world to build a dataset. Additionally, these datasets may be limited to a small number of captured scenes and devices, making it challenging to apply the results to other situations. Therefore, it is common to synthesize degraded images from clean images as a strategy. Most existing image restoration datasets are designed for specific degradation types, such as DIV2K~\cite{DIV2K} and Flickr2K~\cite{Flickr2K}, which are commonly used for SR tasks and include 800 and 2,650 high-resolution images, respectively. For denoising and dejpeg tasks, there are 400 images in the BSDS dataset~\cite{BSD400} and 4,744 images in Waterloo Exploration Database (WED) dataset~\cite{WED}. Some color denoising models  are trained on the polyU Real World Noisy Images Dataset~\cite{polyU} and Smartphone Image Denoising Dataset (SIDD)~\cite{sidd}. For deraining tasks, Rain100L/Rain100H~\cite{rain100}, Rain800~\cite{rain800}, Rain1200~\cite{rain1200}, Rain1400~\cite{rain14000} and Rain13K~\cite{mprnet} are widely used. However, nearly all of these datasets have a limited scale, making it difficult to train large-scale models. ImageNet~\cite{ImageNet}, which consists of about 1.3 million images, is recently used for image restoration pre-training~\cite{IPT,EDT}, but it still suffers from low resolution and high compression ratios.  Recently, the large-scale Laion-5B \cite{schuhmann2021laion} dataset was released for multimodal pretraining, and one intuitive way to create a large-scale restoration dataset is to use the high-resolution subset of Laion-5B, called ``Laion-HR" .  However, most images in Laion-HR lack rich texture details, which are essential for low-level image restoration tasks. In the following section, we will provide a detailed analysis of existing datasets from five aspects and show that our dataset is the first to meet all the requirements.

  \begin{figure*}[!t]
    \centering
      \includegraphics[width=0.9\textwidth]{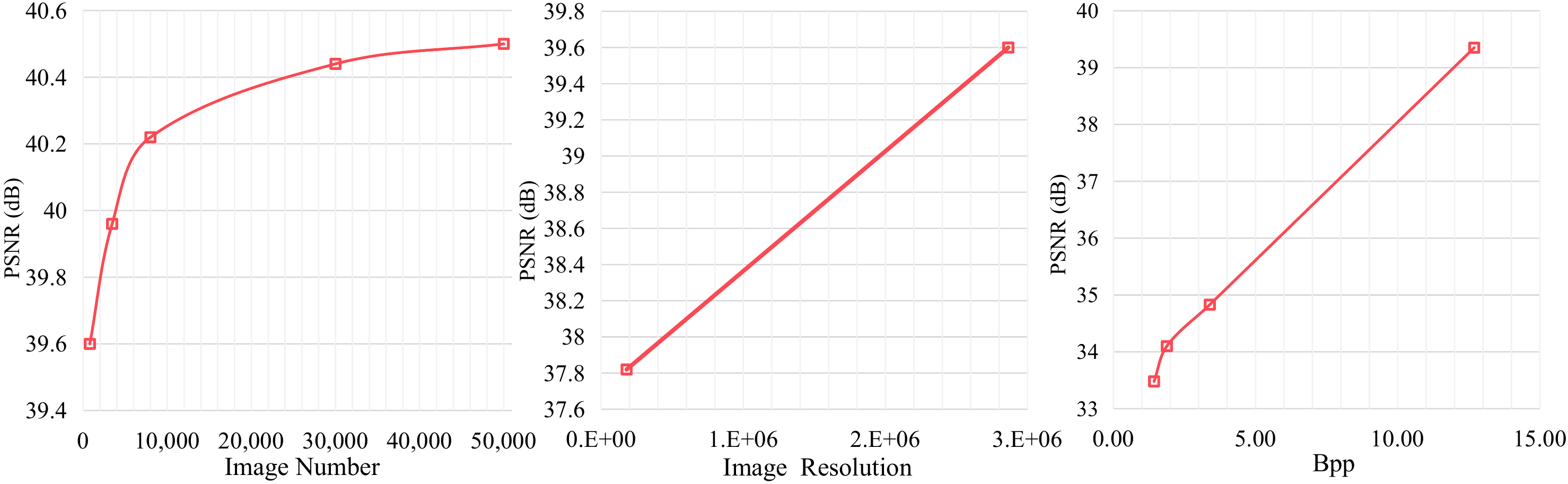}
      \caption{Study on different aspects of dataset on model performance ((a) image number, (b) image resolution , (c) Bpp). Results are tested on Manga109 \cite{manga109} for image SR (×2).}
      \label{fig:data dimension}
  \end{figure*}

\vspace{0.5em}
\noindent\textbf{Image Restoration Methods.}
In recent years, the rapid development of deep learning has led to impressive performance in data-driven methods for image restoration tasks, such as denoising \cite{RIDNet,VDN,chen2020controllable}, super-resolution \cite{RCAN,SAN,HAN}, dejpeg \cite{li2017efficient,yeh2021deep}, and deraining \cite{chen2019gated,MSPFN,PreNet}. At an early stage, CNN-based methods were trained for a specific degradation type with a fixed degradation level. Taking SR as an example, deep learning based methods~\cite{SRCNN,ESPCN,VDSR,LAPSRN,RCAN,SAN,HAN} achieved amazing results, but the model was designed solely for the SR task and could only handle a specific down-sampling scale. This poor generalization and limited applicability to other tasks also existed in deraining~\cite{chen2019gated,DerainNet,derain1,semide_rain} and denoising~\cite{BSD432,ffdnet,chen2020controllable}.
Recently, more methods have shown impressive performance on various image restoration tasks. Among them, transformer-based methods~\cite{IPT,EDT,Uformer,Restormer,swinir} are usually superior to CNN-based methods~\cite{simple,hinet,maxim,MIRNet}, thanks to their strong ability to learn long-range dependencies between image patch sequences. While these models are suitable for multiple tasks, they still need to be trained separately for each specific task and degradation level in order to achieve good performance.

Recently, some latest research has been proposed to study unified image restoration, aiming to restore images under various degradation in a single network. For instance, AirNet~\cite{allinone} proposed to extract specific degradation types through a learned encoder, and GIQE~\cite{shyam2022giqe} randomly combined various degradation types during training. To tackle multiple types of adverse weathers, Chen \etal. ~\cite{learning_multiple} proposed a unified model with two-stage knowledge learning and TransWeather~\cite{transweather} proposed a transformer-based end-to-end model. In this paper, inspired by the MoE idea, we propose a MoE-based unified image restoration model that can handle different low-level tasks including SR, denoising, dejpeg and derain. Although MoE models have been extensively studied in scaling up both vision~\cite{v-moe,wu2022residual} and text models~\cite{gshard,moe2017,switch}, to the best of our knowledge, this is the first attempt at applying MoE to low-level vision tasks.

\section{\dataset~Dataset}
\label{sec:data}

In this section, before introducing our \dataset, we first present a comprehensive analysis of the existing restoration datasets based on the five aforementioned aspects. 

\vspace{0.3em}
\noindent\textbf{Large-Scale.} Large-scale datasets have been proven to be crucial for achieving good results in deep learning for image understanding, particularly for training large models. In order to demonstrate the significance of data scale in the image restoration field, we trained SwinIR~\cite{swinir} on different subsets of our \dataset (which will be described later), each with a different number of training images but the same attributes across four other aspects. As shown in the left of Figure \ref{fig:data dimension}, the performance of SwinIR improves as the number of training images increases, showing that larger-scale dataset also plays an essential role in image restoration field. Nevertheless, existing restoration datasets that have been specifically collected remain severely limited in scale.

\vspace{0.3em}
\noindent\textbf{High Resolution.} Despite some existing restoration works \cite{IPT,EDT} utilizing ImageNet \cite{ImageNet} for large-scale pretraining, they still require further fine-tuning on specific high-quality restoration datasets. This is partially due to the fact that most images in ImageNet have low resolution and quality. In order to systematically explore the impact of resolution, we resize the images from DIV2K \cite{DIV2K} to the same resolution as the average pixel count of images in ImageNet (200,000 pixels). As depicted in the middle section of Figure \ref{fig:data dimension}, decreasing the image resolution significantly impairs the performance of SR. 

\vspace{0.3em}
\noindent\textbf{Low Compression Rate.} The majority of images in existing restoration datasets are sourced from the internet, and many of them have been compressed using image compression techniques to reduce storage requirements. However, high compression rates can lead to significant information loss and distortion artifacts. Since the goal of image restoration is often to generate high-quality images, training models on highly compressed images will result in poor performance. To quantitatively assess the impact of compression ratios on performance, we trained SwinIR on the DIV2K dataset using different levels of compression, as measured in bits per pixel (bpp). A higher bpp value indicates a lower degree of compression for an image. As illustrated on the right-hand side of Figure \ref{fig:data dimension}, lower compression ratios enable the model to achieve better performance.

\begin{figure}
    \centering
    \includegraphics[width=0.45\textwidth]{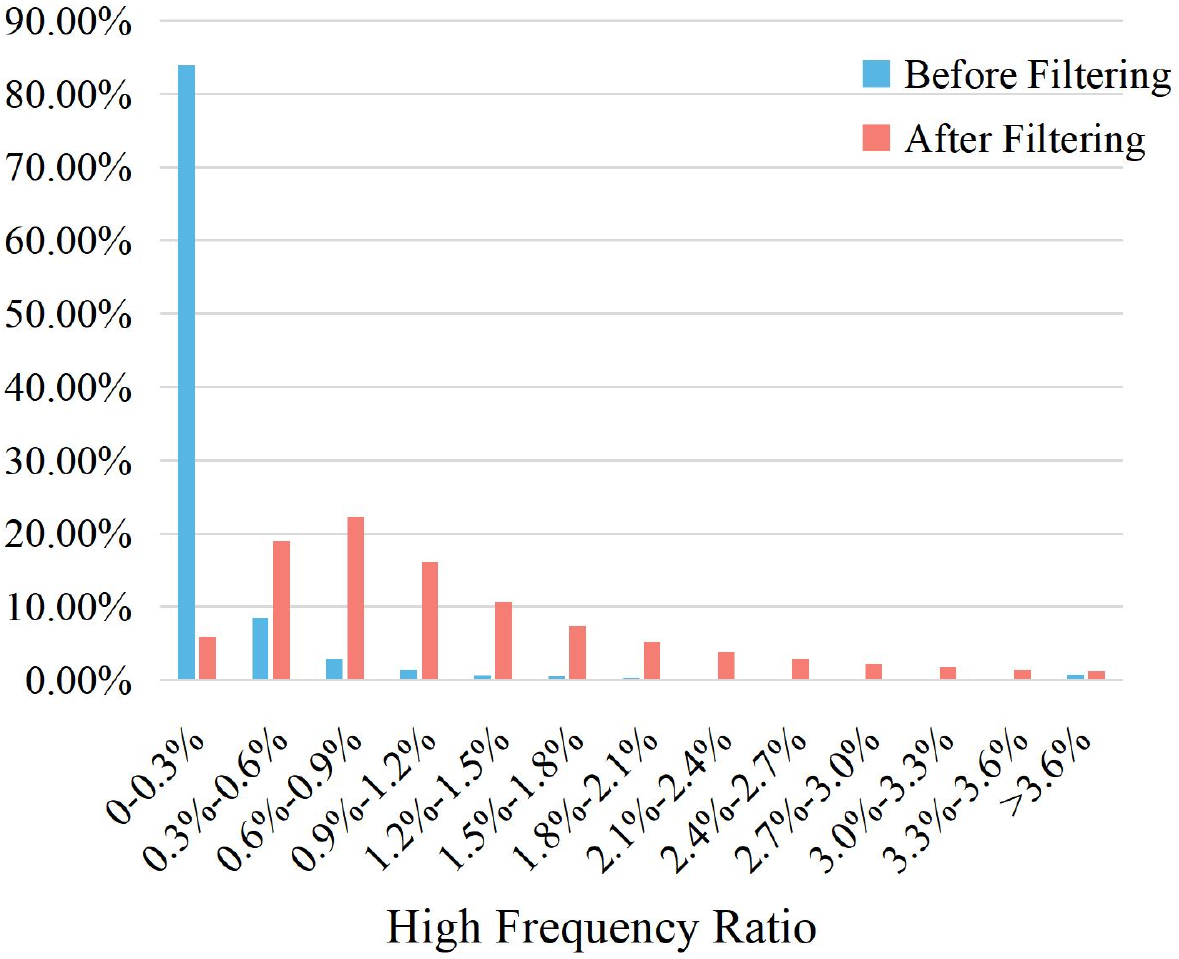}
    \caption{The frequency distribution of the original Laion-HR (blue histogram) and our \dataset~ (orange histogram).}
    \label{fig:freq_dis}
\end{figure}

\vspace{0.3em}
\noindent\textbf{Rich Texture Details.} Similar to compression rates, the richness of image texture details in a dataset would also significantly impact model performance. Take the high-resolution subset of Laion-5B \cite{schuhmann2021laion} (referred to as ``Laion-HR") as a counter-example, although the images in this subset have high resolutions and a large data scale, its texture information details are seriously insufficient to train a good restoration model. As shown in Figure \ref{fig:freq_dis}, frequency analysis \footnote{Here we map the image into the frequency domain with DFT transformation, then calculate the ratio of high-frequency component (i.e., High-frequency ratio) in the whole image.} performed on Laion-HR shows that over 80\% of the images in this subset lack sufficient high-frequency information.
This makes the model trained on Laion-HR incapable of handling images with complex texture structures that contain high-frequency information, such as animal fur, leaves or branches, dense windows in buildings, and other fine-scale regular structures. To validate this, we trained SwinIR on both DIV2K and Laion-HR and compared their performance on Urban100 \cite{urban100}, as shown in Table \ref{table:freq_classify}. The model trained on Laion-HR performs considerably worse than the model trained on DIV2K. We further divide Urban100 into two subsets: high-frequency images and low-frequency images, and find that larger performance gap exists in high-frequency images. 

\begin{table}[!t]
\small
  \caption{Results on different frequency bands of Urban100~\cite{urban100}}
  \centering
  \label{table:freq_classify}
  \setlength{\tabcolsep}{0.5pt}{
\begin{tabular}{l|c|c|c}
\hline
        & Urban100 & Low-freq. Subset       & High-freq. Subset      \\
         \hline
High-freq Ratio& 1.514\%& 0.2035\%& 15.84\% \\
DIV2K   & 33.34/0.9391& 37.99/0.9625& 32.29/0.9359       \\
Laion-HR& 32.64/0.9324& 37.43/0.9586 & 31.43/0.9280   \\
\hline 
\end{tabular}}
\end{table}

\noindent\textbf{Semantic Coverage.} To train a generalizable model that performs well on images of different semantic categories, in our work, we categorize images into three broad classes: outdoor, indoor, and artificial. The outdoor class contains images captured in nature and architecture, people, animals, and transportation. The indoor class includes images of indoor scenes, furniture, food, and objects. The artificial class contains images that have been post-processed or synthesized directly, such as posters, maps, texts, and comics. To measure the diversity of category distribution in a dataset, we calculate the entropy of semantic category using the formula: $H =-\sum_{i=1}^{n}p_i\log p_i$ where $p_i$ is the percentage of the $i$-th class in the total dataset. The larger the value of $H$, the more evenly distributed the dataset is across all the semantic categories we define.

\begin{figure}
    \centering
    \includegraphics[width=0.43\textwidth]{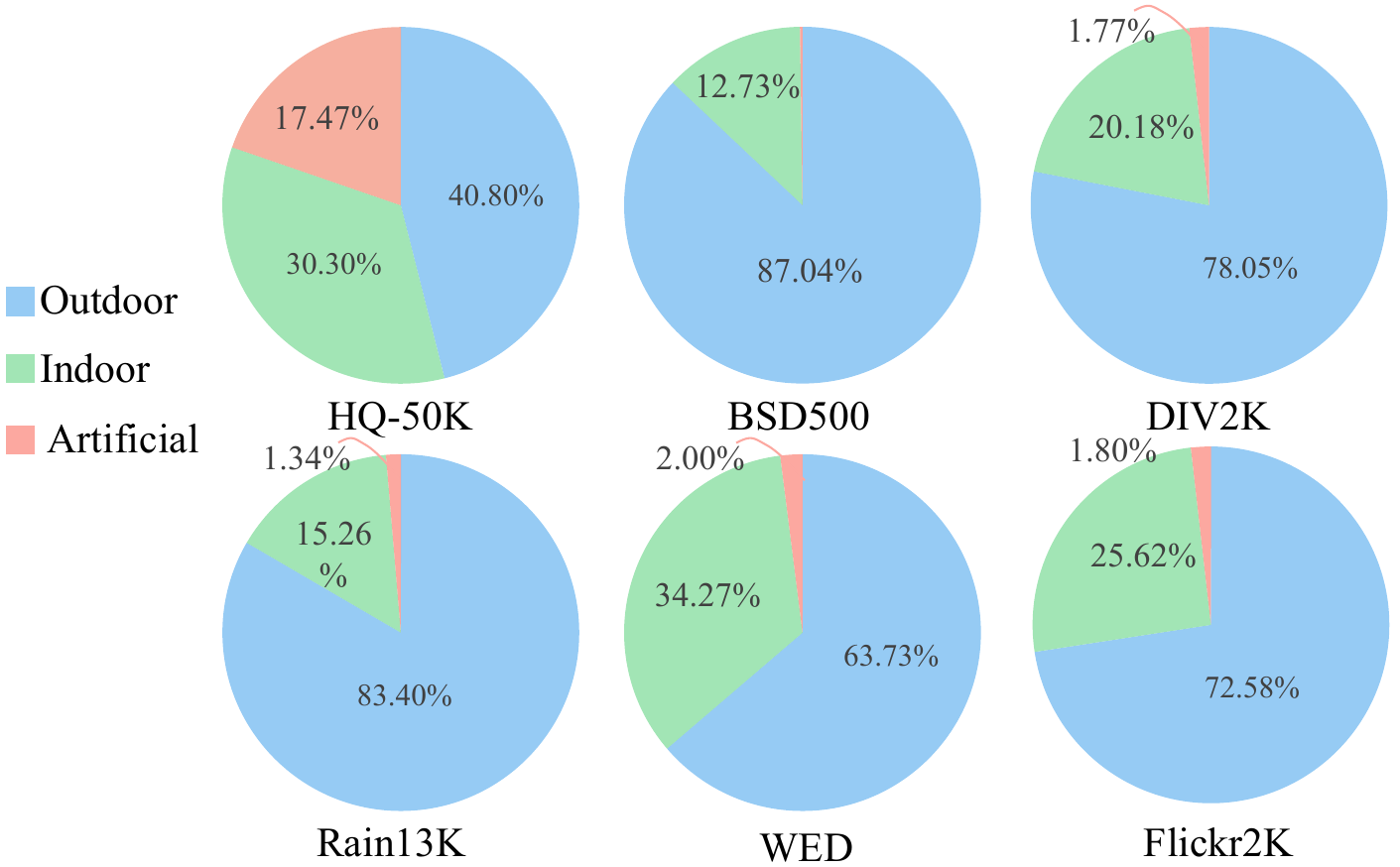}
    \caption{Semantic distribution of existing datasets in three broad categories: outdoor, indoor, artificial.}
    \label{fig:semantic_dis_bigclass}
\end{figure}

Our analysis of existing image restoration datasets revealed that they often have an unbalanced semantic distribution (shown in Figure \ref{fig:semantic_dis_bigclass}), with a large portion of images belonging to the outdoor class, while indoor and artificial images are relatively fewer (especially artificial images). Our experiments (Table \ref{table:sr_impro_semantic} and supplementary material) demonstrate that training a model on such an unbalanced dataset results in suboptimal performance on categories with limited coverage.

\vspace{0.3em}

\begin{table*}[!t]
\small
  \centering
    \caption{Statistical comparison of different datasets across five aspects: Scale, Resolution (Avg.pixels), Compression rates(bpp), Texture details (High-frequency ratio), and Semantic Coverage (Diversity).}
    \label{table:dataset_statistics}
    \setlength{\tabcolsep}{10pt}{
  \begin{tabular}{l|cccccc}
  \hline
    Dataset  & Task              & Scale$\uparrow$     & Avg.pixels$\uparrow$ & bpp $\uparrow $           & \begin{tabular}[c]{@{}c@{}}High-freqency Ratio\end{tabular} $\uparrow $ & Diversity $\uparrow$\\\hline
    DIV2K~\cite{DIV2K}    & \begin{tabular}[c]{@{}c@{}}SR\end{tabular}      & 800       & 2,788,027  & 12.69           & 0.8227\%                                &       0.997            \\
    Flickr2K~\cite{Flickr2K} &\begin{tabular}[c]{@{}c@{}}SR\end{tabular}    & 2,650      & 2,763,878    & 12.76         & 0.5369\%                                & 1.032                        \\
    BSDS400~\cite{BSD400}  & \begin{tabular}[c]{@{}c@{}}Denoising\end{tabular}      & 400       & 154,401     & 3.77          & 0.8863\%                                & 1.092 
                            \\
    WED~\cite{WED}      & \begin{tabular}[c]{@{}c@{}}Denoising, Dejpeg\end{tabular}         & 4,744      & 218,576     & \textbf{24.74}        & 0.5818\%                                & 1.068 
                             \\
    Rain13K~\cite{mprnet}  & Derain                                                                       & 1,212    &  209,256
   &  1.5357
       &    1.0858\%
                           & 0.934
                          \\\hline
    ImageNet &\begin{tabular}[c]{@{}c@{}}High-level Tasks\end{tabular}                   & 1,281,167   & 219,175     & 4.72        & 1.0930\%                                & 1.036                          \\
    Laion-HR &\begin{tabular}[c]{@{}c@{}}Multimodal Pretraining\end{tabular}                   & \textbf{170,000,000} & \textbf{3,117,066}   & 1.57           & 0.1918\%                                & 1.099 
                              \\
    \dataset~(Ours)   & \begin{tabular}[c]{@{}c@{}}Low-level Tasks\end{tabular}   & 50,000     & 2,509,509    & 12.86       & \textbf{1.4270\%}                              & \textbf{1.143} \\
\hline                    
  \end{tabular}}
  \end{table*}

\begin{figure*}[h]
  \centering
  \includegraphics[width=\linewidth]{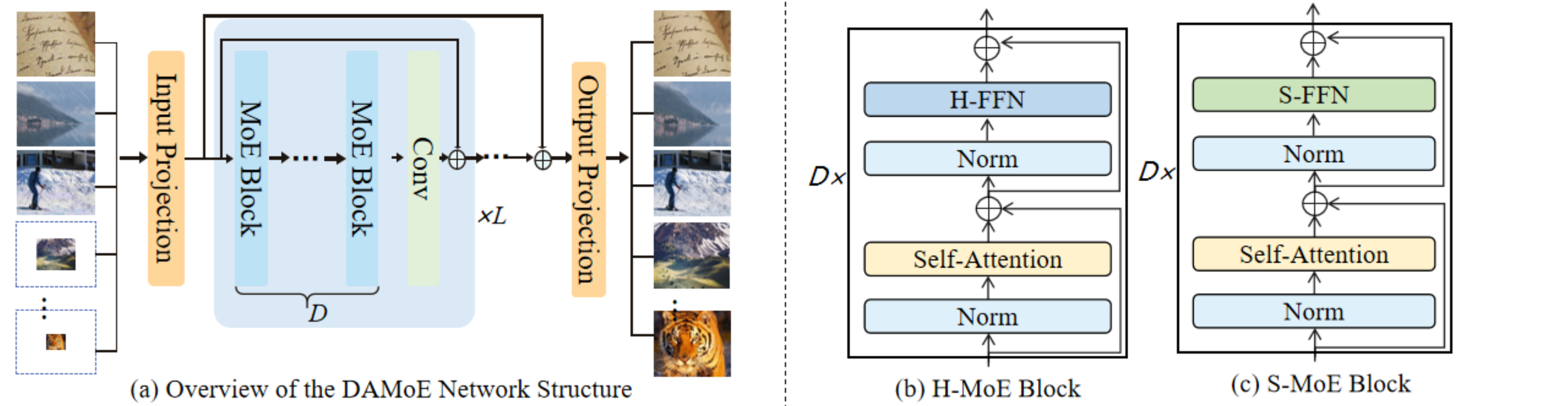}
  \caption{Illustration of the structure of Degradation-Aware Mixture of Expert (DAMoE). $\oplus$: element-wise addition. (a) Overview of DAMoE; (b) H-MoE Block introduces H-MoE layer while (c) S-MoE Block introduces S-MoE layer.} 
  \label{fig:backbone}
\label{fig:module}
\end{figure*}

\subsection{Dataset Collection and Processing of \dataset}
\label{filter}
To ensure the high quality of the proposed dataset, we have taken into account the aforementioned five aspects during the construction process. First, we collect a large number of images from the Internet and existing large-scale dataset~\cite{schuhmann2021laion}. To ensure \textit{balanced semantic coverage}, we select images from the above three broad categories, each of which contains several sub-categories. Since we do not have the semantic label for the collected images, we instead leverage the pretrained CLIP \cite{radford2021learning} to categorize each image. Moreover, we try to balance the image number of each category and guarantee that each category would have a sufficient number of images. The final semantic distribution can be seen in Figure \ref{fig:semantic_dis_bigclass}. 

Next, we apply a series of filtering strategies to carefully remove images that did not meet the requirements. Specifically, to fulfill the \textit{high resolution} requirement, we discard images with a length or width smaller than 1024 pixels. The final average pixel number is 2,509,509. To meet the \textit{low compression ratio} requirement, we filter out the images that occupy less than 500KB storage space,  resulting in a final average bpp of \dataset~at 12.86. For the \textit{texture detail} requirement, we calculate the high-frequency ratio of each image and removed images with a high-frequency ratio of less than 0.5\%. We also ensure that the distribution of high-frequency ratio was wide enough to approach a normal distribution, thereby improving the generalization of each frequency domain interval. After filtering, the final frequency distribution is displayed in Figure \ref{fig:freq_dis} (orange histogram). Ultimately, 50,000 high-quality images, which is approximately 5.6\% of all collected images, are selected and denoted as \dataset. 

In addition to the 50,000 training images, we further collect 1250 test images that spanned across each semantic sub-category to help analyze the performance in a fine-grained manner. 

\subsection{Data Statistics of \dataset~}
Table \ref{table:dataset_statistics} presents the detailed statistics of our \dataset~ and existing image restoration datasets. To summarize, \dataset~ is superior to existing datasets in terms of:

\vspace{0.3em}
\noindent\textbf{Number of images.} To the best of our knowledge, \dataset~ has the largest scale compared with other datasets that are dedicated for image restoration tasks. It contains 50K images, which is 60$\times$ larger than DIV2K\cite{DIV2K}.

\vspace{0.3em}
\noindent\textbf{High resolution.} The average pixel number of each image in \dataset~ is comparable to DIV2K\cite{DIV2K} and Flickr2K\cite{Flickr2K}, but much larger than BSDS400\cite{BSD400}, WED\cite{WED}, Rain13K and ImageNet\cite{ImageNet}.

\vspace{0.3em}
\noindent\textbf{Low compression ratio.} \dataset~ is carefully filtered as stated in Section \ref{filter}. It takes the second place in terms of bpp, which is only inferior WED~\cite{WED}. However, \dataset~ has 10$\times$ more images than WED~\cite{WED}

\vspace{0.3em}
\noindent\textbf{Rich texture Details.} The proposed \dataset~ has the best texture detail richness as measured by high-frequency ratio. Compared to Laion-HR, it also has a more balanced frequency distribution as shown in Figure \ref{fig:freq_dis}.

\vspace{0.3em}
\noindent\textbf{Semantic diversity.} \dataset~ also show the best semantic coverage and most balanced semantic distribution as shown in Figure \ref{fig:semantic_dis_bigclass}. We have visualized the example images for each semantic sub-category in the supplementary material.  

\section{DAMoE for Unified Image Restoration}
Our DAMoE is designed to handle a wide range of restoration tasks (\eg, denoising, SR, deraining and dejpeg) and degradation levels (\eg, $\times 2, \times 3, \times 4$ settings in SR) through a unified single model. As illustrated in Figure \ref{fig:module} (a), DAMoE employs shared modules and parameters for different tasks with different levels of degradation, except for the experts in MoE layers, which are dedicated to specific tasks (H-MoE) or dynamically activated based on input (S-MoE).
Specifically, let $H\times W$ denote the input resolution of DAMoE. For tasks with lower resolution inputs (e.g., SR), we standardize the input resolution by upsampling the input to $H \times W$ with bicubic interpolation. Then, for an input image $I_{in} \in \mathbb{R}^{3 \times H \times W}$,  a convolution-based projection layer is utilized to extract shallow features $X \in \mathbb{R}^{C\times H\times W}$ from $I_{in}$, where $C$ represents the feature channel dimension. 
Before feeding $X$ to several following transformer blocks, a learnable task embedding $e^{task}_{i} \in \mathbb{R}^{C\times 1 \times 1}$, $i =\left\{1,...,N\right\}$, where $N$ indicates the number of tasks, is added to $X$.
The backbone mainly comprises multiple MoE blocks and convolution layers. Thanks to the effective receptive fields of shifted windows~\cite{SWINTRANSFORMER} and self-attention, high-frequency texture and structural details of images can be effectively restored. We also use several skip-connections to combine shallow features with deep features. Finally, the output image $I_{out}\in \mathbb{R}^{3\times H\times W}$ is reconstructed by a convolution-based projection layer.

\begin{figure}
    \centering
    \includegraphics[width=\linewidth]{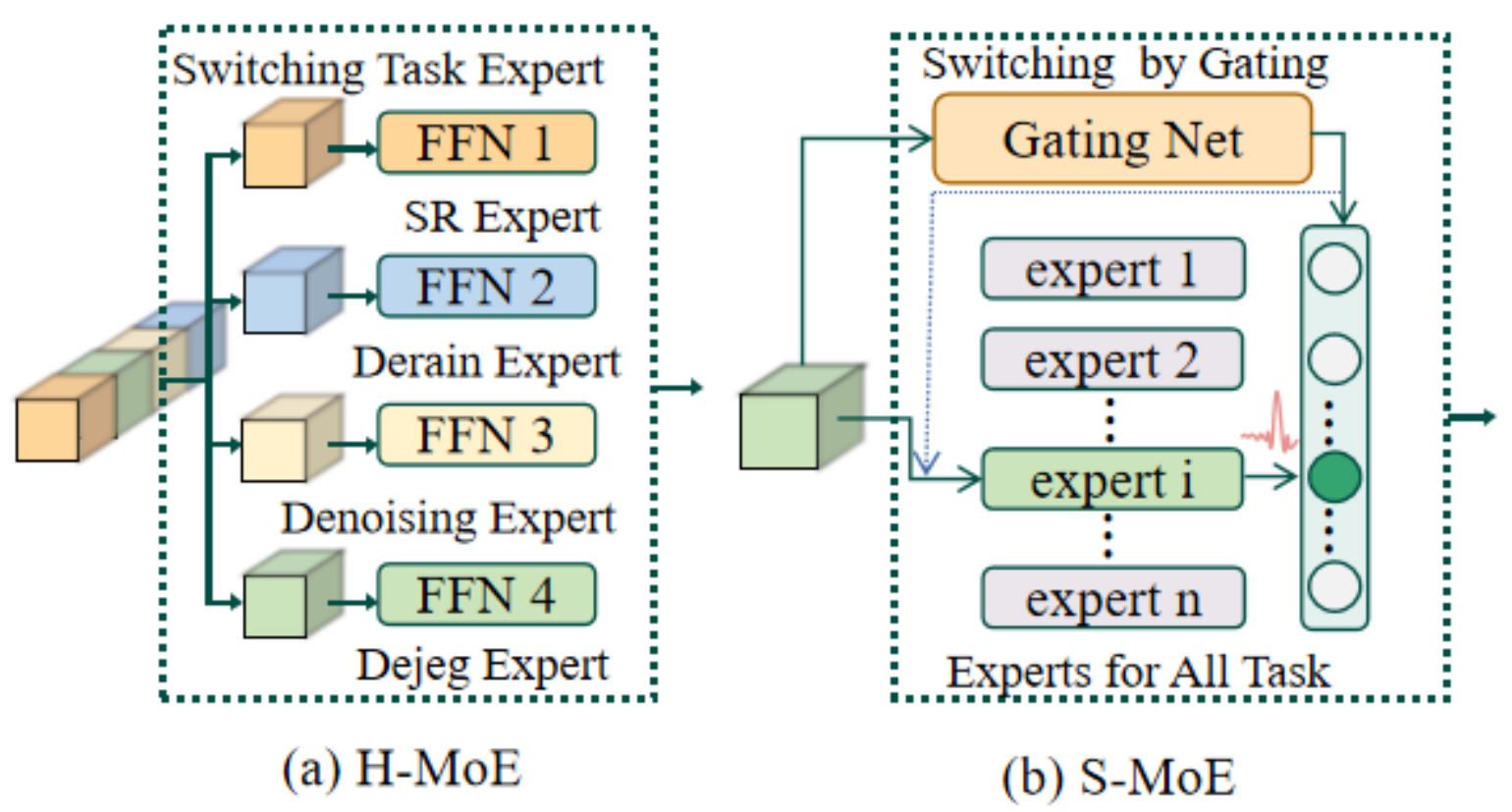}
    \caption{(a) H-MoE layer applies task-specific experts to different tasks. For one task, different levels of degradation share the same expert. (b) S-MoE layer processes multi-task input by a gating network, which means that all experts are visible to all tasks.}
    \label{fig:moe_block}
\end{figure}

\setlength{\tabcolsep}{4pt}
\begin{table}[!t]
\small
  \caption{Super resolution results on Set14 and Manga109 by using different training datasets.}
  \centering
  \label{table:sr_impro}
  \setlength{\tabcolsep}{6.5pt}{
\begin{tabular}{c|c|cc|cc}\hline
 \multirow{2}{*}{scale} & \multirow{2}{*}{Training Set} & \multicolumn{2}{c|}{Set14} & \multicolumn{2}{c}{Manga109} \\
                                               & \multicolumn{1}{c|}{}                              & PSNR        & SSIM        & PSNR         & SSIM          \\\hline
 \multirow{2}{*}{x2}    & DF2K                                              & 34.46       & 0.9250      & 39.92        & 0.9797        \\
                          & HQ-50K                                            & \textbf{34.81}       & \textbf{0.9262}      & \textbf{40.50}         & \textbf{0.9811}       \\ \cline{1-6}
                        \multirow{2}{*}{x3}    & DF2K                                              & 30.93       & 0.8534 & 35.12        & 0.9537        \\
              & HQ-50K                                            & \textbf{31.15}       &    \textbf{0.8542}    & \textbf{35.21}     &   \textbf{0.9541}       \\\cline{1-6}
                        \multirow{2}{*}{x4}    & DF2K                                              & 29.09       & 0.7950       & 32.03        & 0.9260        \\
                                          & HQ-50K                                            & \textbf{29.35}       & \textbf{0.7953}     & \textbf{32.36}        & \textbf{0.9277} \\ \hline     
\end{tabular}  }
\end{table}
\setlength{\tabcolsep}{4pt}

\subsection{Mixture of Expert Layer}
\label{sec:moe}
Taking inspiration from the Mixture of Expert (MoE) mechanism~\cite{moe_first,moe2017,switch}, we incorporate MoE layers in our transformer backbone by substituting the feed forward network (FFN) layer with a MoE layer. The MoE layer comprises a collection of $n$ experts {$E_1$, $E_2$, ..., $E_n$} and a routing strategy. Given an input feature $x \in \mathbb{R}^{C}$, which represents a feature vector from a single spatial position of the feature maps,  the MoE layer produces the output $y \in \mathbb{R}^C$, which is the weighted summation of different experts: 
\begin{equation}
\label{equation:moe}
    y=\sum_i^n{w(x)_i E_i(x)},
\end{equation}
where $w(x) \in \mathbb{R}^n$ is the weight of different experts for the input $x$. In this paper, we design two types of MoE layers, named as H-MoE and S-MoE, as illustrated in Figure \ref{fig:backbone} (b) and (c). The key difference between the two is in their routing strategy. For H-MoE (Hard-MoE), $w(x)$ is represented as a one-hot vector. This means that the input feature $x$ is only visible to the corresponding task-specific expert. In contrast, for S-MoE (Soft-MoE), $w(x)$ is obtained through a gating network:
\begin{equation}
    w(x) = \mathcal{Z}_k(\textrm{softmax}(\mathbf{W_g}\otimes x_i)),
    \label{eq:weight of moe}
\end{equation}
where $\mathbf{W_g} \in \mathbb{R}^{n \times C}$ is a trainable parameter, $\otimes$ is the matrix multiplication, $\mathcal{Z}_k(\cdot)$ is the function that sets all values to zeros except the top-$k$ largest values.

\begin{table}[!t]
\small
  \caption{Super resolution results on our proposed fine-grained evaluation benchmark by using different training datasets.}
  \centering
  \label{table:sr_impro_semantic}
  \setlength{\tabcolsep}{2pt}{
\begin{tabular}{c|c|cc|cc|cc}\hline
 \multirow{2}{*}{scale} & \multirow{2}{*}{Training Set} & \multicolumn{2}{c|}{Outdoor} & \multicolumn{2}{c|}{Indoor}  & \multicolumn{2}{c}{Artificial}\\
                                               & \multicolumn{1}{c|}{}                              & PSNR        & SSIM        & PSNR         & SSIM  & PSNR         & SSIM          \\\hline
   \multirow{2}{*}{x2}    & DIV2K                                              &  30.77    &  0.9008    &   31.21
   &0.9053   &  30.31	&  0.9120     \\
                          & HQ-50K                                            &  \textbf{31.00}    & \textbf{0.9067}     &   \textbf{31.52}     &  \textbf{0.9134} &  \textbf{30.46} & \textbf{0.9257} \\ 
            \cline{1-8}
   \multirow{2}{*}{x3}    & DIV2K                                              &   27.54  &  0.8044    &    28.00    & 0.8201   & 26.64  &  0.8384     \\
                         & HQ-50K                                            &  \textbf{27.69}     &   \textbf{0.8086}  &     \textbf{28.20}   &  \textbf{0.8224}   &  \textbf{27.47} &   \textbf{0.8504} \\
                         \cline{1-8}
                        \multirow{2}{*}{x4}    & DIV2K                        &  25.97     &   0.7334  &    26.40     &   0.7557 &  24.79 &   0.7799    \\
                                          & HQ-50K                           &  \textbf{26.18}     & \textbf{0.7388}    &  \textbf{26.60}       &  \textbf{0.7614} & \textbf{25.81}  & \textbf{0.7862}\\ \hline     
\end{tabular}  }
\end{table}

\begin{table}[!t]
  \small
  \caption{Results comparison (``PSNR" ) on the denoising task by using different training datasets. }
  \centering
  \label{table:denoise_impro}
\setlength{\tabcolsep}{1.0mm}{
  \begin{tabular}{c|lll|lll}
 \hline
  \multirow{2}{*}{TrainingSet}  & \multicolumn{3}{c|}{CBSD68\cite{CBSD68}}             & \multicolumn{3}{c}{Kodak24\cite{Kodak24}}   \\ \cline{2-7} 
                 &      15    & 25    & 50 & 15       & 25      & 50                \\ \hline
 BSDS400+WED         & 34.29 & 31.66 & 28.43           &35.25 &32.82 &29.79\\
  \dataset          &  \textbf{34.38} & \textbf{31.70}    &    \textbf{28.53 }                    &   \textbf{35.30 }      &    \textbf{ 32.83}    &   \textbf{30.03}       \\\hline
  \end{tabular}}
  \end{table}

\begin{table}
    \centering
    \small
    \caption{Results comparison (``PSNR" ) on the Dejpeg task by using different training datasets.  }
    \label{table:jpeg} 
    \vspace{3pt}
\setlength{\tabcolsep}{3.2mm}{
  \begin{tabular}{c|llll}
 \hline
  \multirow{2}{*}{TrainingSet}  & \multicolumn{4}{c}{Classic5\cite{classical5}}             \\ \cline{2-5} 
                 &      10    & 20    & 30 & 40                    \\ \hline
 BSDS400+WED         & 30.27 & 32.25 & 33.69           &34.52\\
  \dataset           &  \textbf{30.34}    &   \textbf{32.44}   &     \textbf{33.72}                    &   \textbf{34.56}       \\\hline
  \end{tabular}}
\end{table}

  \begin{table}
  \small
    \caption{Results comparison on the deraining task by using different training datasets.}
    \centering
    \label{table:derain_impro}  
    \setlength{\tabcolsep}{1.2mm}{
    \begin{tabular}{c|c|ll|cc}
   \hline 
    \multirow{2}{*}{Method} & \multirow{2}{*}{TrainingSet} & \multicolumn{2}{c|}{Rain100L\cite{rain100}}  & \multicolumn{2}{c}{Rain100H\cite{rain100}}  \\\cline{3-6}
                                  &                             & PSNR    & SSIM    & PSNR & SSIM       \\ \hline
    Restormer\cite{Restormer}         & Rain13K    &  38.99      & 0.9780  &  31.46 &   0.9040  \\ \hline
    SwinIR\cite{swinir}         & Rain13K    &  35.97       & 0.9640  & 29.80  &  0.8827    \\ \hline
    SwinIR\cite{swinir}         & \dataset     &  \textbf{41.86}	   & \textbf{0.9894}     & \textbf{32.52} & \textbf{0.9257}                    \\\hline
    \end{tabular}}
  \end{table}

\begin{table*}[h]
  \caption{Multi-task results using a unified model trained on both our \dataset~ and the combination of existing restoration datasets from all the tasks involved. The best results are in bold.}
  \centering
  \label{table:multask}
  \small
\setlength{\tabcolsep}{1.7mm}{
  \begin{tabular}{l|l|ccc|c|ccc|cccc}
 \hline
  \multirow{3}{*}{Training Set} & \multirow{3}{*}{Method} & \multicolumn{3}{c|}{SR (PSNR-Y)}      & \multicolumn{1}{c|}{DeRain (PSNR-Y)}   & \multicolumn{3}{c|}{Denoise (PSNR)} & \multicolumn{4}{c}{Dejpeg (PSNR)} \\
                                &                         & \multicolumn{3}{c|}{Manga109} & \multicolumn{1}{c|}{Rain100L} & \multicolumn{3}{c|}{CBSD68}  & \multicolumn{4}{c}{Classic5}  \\\cline{3-13}
                                &                         & x2       & x3      & x4     &      -    & 15      & 25      & 50      & 40    & 30   & 20   & 10   \\\hline
   \multirow{4}{*}{Combination}                            & AirNet\cite{allinone}                             &  36.98       &    25.02      &    23.24     &   31.20    &      33.50    &    30.67     &    26.35     &    33.73    &    32.88   &    31.58  &  29.25        \\
                              & SwinIR\cite{swinir}                             &  36.72  &  31.52       &  25.78      &  36.28     &33.82          &31.19         & 27.89        & 33.77      & 32.91    &30.75      & 29.39         \\
                             & DAMoE                                       &    \textbf{37.95}               &     \textbf{32.93}          &    \textbf{25.89}         &  \textbf{38.04}                &        \textbf{34.12}        &      \textbf{31.47}           &       \textbf{28.22}       &    \textbf{34.00}   &   \textbf{33.16}   &   \textbf{31.87}   &  \textbf{29.67}    \\
\hline
   \multirow{4}{*}{\dataset~ }                                              & AirNet\cite{allinone}                                        &  37.13        &    25.11    & 23.37      &      35.31   &       33.63 &    30.78    &   26.53    &  33.86     &   33.01   &   31.75   & 29.54   \\
                                 &SwinIR~\cite{swinir}                                      & 37.56         &  32.48       &    25.87     &      38.08       &      34.05   &    31.38       &   28.09      &    33.94    &     33.10  &   31.83   &    29.61   \\
                                              & DAMoE                                       &  \textbf{38.51}        &   \textbf{33.40}    &  \textbf{26.12}      &  \textbf{40.22}       &  \textbf{34.18}      &  \textbf{31.52}     &    \textbf{28.28}     &  \textbf{34.15}    &   \textbf{33.31}  &   \textbf{32.03}   &   \textbf{29.80} 
  \\

 \hline
  \end{tabular}
  }
  \end{table*}

\subsection{Mixture of Expert Block}
Equipped with the H-MoE and S-MoE layers described above, we design two types of MoE blocks accordingly. As shown in Figure \ref{fig:moe_block} (a) and (b), we replace  the FFN layer in the original transformer block with these two types of MoE layers. In the \textit{H-MoE block},  the number of experts is the same as the number of tasks, with each expert responsible for one task. Different tasks have no mutual influence within the H-MoE layer. Note that we share the same expert for different degradation levels of one task. In the \textit{S-MoE block}, the number of experts in the S-MoE layer is a hyper-parameter and does not need to be the same as the number of tasks. The S-MoE block learns how to use different combinations of experts for different tasks and degradation levels. DAMoE can use one type of MoE block or a combination of both by alternating between the H-MoE and S-MoE blocks. A detailed performance analysis will be provided in the ablation analysis section.

\section{Experiments}
\label{sec:experiements}
\noindent\textbf{Experiment Setup.}
We conduct experiments on four representative image restoration tasks including SR, deraining, dejpeg and denoising. We train the models on both our \dataset~ and existing restoration datasets, and compare the performance. As the baseline, for SR, we use the DF2K dataset, which is a combination of DIV2K \cite{DIV2K} and Flickr2K \cite{Flickr2K}). For deraining, we use the Rain13K dataset, which is a combination of Rain14000~\cite{rain14000}, Rain1800~\cite{rain1800}, Rain800~\cite{rain800}, and Rain12~\cite{Rain12}. For denoising and dejpeg, we use the combination of BSD400~\cite{BSD400} and WED~\cite{WED} as the training set.

To create training image pairs, we follow the standard setting\cite{allinone,swinir,Restormer} and consider different degradations as follows: 1) For SR, bicubic is used to obtain downsampled image; 2) For denoising, we add white Gaussian noise with $\sigma=\left\{15,25,50\right\}$ to the color images; 3) For dejpeg, we compress images with JPEG quality $q=\left\{10,20,30,40\right\}$; 4) For deraining, we use the synthetic rain method~\cite{rain100} to create the corresponding raining image. 

  \begin{table}
    \centering
    \caption{Ablating the four aspects in dataset requirements on $\times2$ SR task. The ablated aspect in each row is marked with blue color.}
\label{tab:five dim}
\small
\setlength{\tabcolsep}{2.5mm}{
\begin{tabular}{c|c|c|c|ll}
\hline
 \multirow{2}{*}{Res.} & \multirow{2}{*}{H-freq} & \multirow{2}{*}{Bpp} & \multirow{2}{*}{Div.} & \multicolumn{2}{c}{Manga109\cite{manga109}} \\\cline{5-6}
                                                       &                                  &                      &                          & PSNR         & SSIM          \\ \hline
\color{blue}{279,948}                                         & 0.0183                                               & 12.34                                    & 1.143                                          & 37.88        & 0.9758       \\                                             
 2,519,300                                       & \color{blue}{0.0011}                                               & 12.86                                    & 1.143                                          & 35.51        & 0.9663        \\
 2,519,300                                       & 0.0134                                               & \color{blue}{1.88}                                     & 1.143                                          & 34.10         & 0.9521        \\

2,513,121                                       & 0.0189                                               & 12.92                                    & \color{blue}{1.001}                                          & 39.17        & 0.9780         \\
 \hline
2,519,300                                       & 0.0185                                               & 12.92                                    & 1.143                                          & \textbf{39.54}        & \textbf{0.9791}        \\\hline
\end{tabular} }
\end{table}

\subsection{Evaluation on Single Task}
We first show the quantitative results of our proposed \dataset~ on each single image restoration task. Here we uniformly use SwinIR~\cite{swinir} as the training framework. Table \ref{table:sr_impro} shows the performance comparison on the SR task, as we can see, significant performance gain can be obtained on both the Set14 and Manga109 dataset by using our \dataset. In Table \ref{table:sr_impro_semantic}, We further present the detailed performance analysis on our proposed fine-grained evaluation benchmark to evaluate the performance on different semantic categories. Due to the limited number of artificial images in DF2K compared to our \dataset~ dataset, which has a more balanced semantic coverage, the model trained on \dataset~ performs significantly better on artificial images. This highlights the importance of semantic coverage and balance when training a generalizable SR model.

For the denoising, dejpeg and deraining tasks, the corresponding results are presented in Table \ref {table:denoise_impro}, Table \ref{table:jpeg} and Table \ref {table:derain_impro} respectively. It can be seen that, our proposed \dataset~ can also achieve consistently better results than the existing restoration datasets or their combination. Especially on the deraining task, the model trained on our \dataset~ outperforms the baseline by a large margin on Rain100L benchmark (i.e., more than 5 dB). 

\begin{table*}[!ht]
\small
  \centering
    \caption{Ablation Study for the effect of different MoE Structure on All-in-One Training.}
  \label{tab:Moe}
\setlength{\tabcolsep}{2.2mm}{
  \begin{tabular}{c|cc|c|c|cc|cc|c|c}
  \hline
  \multirow{3}{*}{}       & \multirow{3}{*}{H-MoE} & \multirow{3}{*}{S-MoE} & \multirow{3}{*}{Params(M)}& \multirow{3}{*}{FLOPs(G)} & \multicolumn{2}{c|}{SR $\times2$}  & \multicolumn{2}{c|}{Derain}    & \multicolumn{1}{c|}{Denoise 15} & \multicolumn{1}{c}{Dejpeg 40} \\
                          &                                                     &                 &                     &                 & \multicolumn{2}{c|}{Manga109~\cite{manga109}} &\multicolumn{2}{c|}{Rain100L~\cite{rain100}}& \multicolumn{1}{c|}{CBSD68~\cite{CBSD68}} & \multicolumn{1}{c}{Classic5~\cite{classical5}}   \\
                          &       &      &    &          &          PSNR & SSIM  &PSNR & SSIM   &PSNR      & PSNR           \\\hline
  \multirow{5}{*}{SwinIR} &         -                 &             -                   &           11.49            &        83.49          &  37.56       &     0.9748            & 38.08    &    0.9798 &    34.05 
      &        33.94                     \\ 
                          &         \checkmark                          &                 &      25.55          &        83.49                  &     37.72     &        0.9754     &   38.12   & 0.9801        &   34.08       &         33.95                 \\
                          &                          &             \checkmark               &      139.05         &        99.98                 &   37.90        &      0.9760   &    38.30  &    0.9809            &      34.11   &     34.03               \\
                          &       \checkmark           &      \checkmark              &       82.30            &       91.74              &     38.51      &       0.9772     &  40.22    &   0.9861         &  34.18        &        34.15                  \\\hline
  \end{tabular}}
  \end{table*}

\begin{table*}[!ht]
\centering
\small
\setlength{\tabcolsep}{4.0mm}{
\caption{The impact of multi-task training on a single model. Multi-task means inputs are in different tasks, while multi-level means inputs with different degradation levels.}
\label{table:single_to_multi_task}
  \begin{tabular}{c|c|c|c|c|c|c}
    \hline
    \multirow{3}{*}{DAMoE} & \multirow{3}{*}{Multi-task} &\multirow{3}{*}{Multi-level} & SR $\times 2$ & DeRain & Denoise $\sigma=15$  & Dejpeg $q=40$                       \\
              &        &         &Manga109 \cite{manga109}  &Rain100L \cite{rain100}  & CBSD68~\cite{CBSD68} & Classic5 \cite{classical5} \\
               &       &         &PSNR/SSIM       & PSNR/SSIM                   &  PNSR          & PSNR   \\\hline
       -    & -            &      -        & 40.05/0.9894     & 41.86/0.9894   & 34.38 & 34.56 \\
     -  &-     &              \checkmark              & 37.91/0.9769   & -  & 34.17 &   34.08       \\
         -  & \checkmark          &         \checkmark      &   37.56/0.9748 & 38.08/0.9798   & 34.05 & 33.94 \\
         \checkmark     &  \checkmark          &         \checkmark      &  38.51/0.9772   & 40.22/0.9860
  & 34.18 &   34.15          \\
         \hline
    \end{tabular}}
\end{table*}

\subsection{Evaluation on Multiple Tasks}
\label{sec:Multaskexp}
In addition to the single task evaluation, we further provide the multi-task evaluation results with existing unified restoration models. Here we choose AirNet~\cite{allinone} and SwinIR~\cite{swinir} as the baselines. In detail, though AirNet~\cite{allinone}~ gave the results of processing different unknown degradation with one unified model, it has not covered the experiments on SR and dejpeg tasks. And SwinIR~\cite{swinir} only provided results of the single task setting to handle a specified task and degradation level. Therefore, we run their officially released code to report the results by following the default settings. We train these baseline models and our DAMoE model on both our \dataset~ and the combination of existing restoration datasets from all the tasks involved. The detailed results are shown in Table~\ref{table:multask}. Comparing the results between using our \dataset~ and the combination of existing datasets, we can observe consistent performance improvement on all different tasks, demonstrating the value of our dataset for training a unified restoration model. On the other hand, when using the same training dataset, our proposed unified model DAMoE also outperforms existing state-of-the-art unified restoration model AirNet \cite{allinone}. This shows that integrating the MoE idea for unifying different restoration tasks is a very promising direction and worthy of further investigation.

\subsection{Ablation Analysis}
\label{sec:Ablation}

\noindent\textbf{Ablation of data requirements.} In Section \ref{sec:data}, we have already provided the detailed analysis of existing datasets from the proposed five aspects. In this section, we further provide the ablation results on our \dataset~. Since the performance trend for different data scales is already given in Figure \ref{fig:data dimension}, we only provide the ablation results on other four aspects in Table \ref{tab:five dim}. Considering the training resource, here we only use a subset of 800 images from \dataset~ for each ablation setting. From the ablation results, we can see that all the four aspects are very important for our \dataset~ and without considering any of each aspect will result in significant performance drop on the SR task.

\noindent\textbf{MoE block design.} As previously described, we have developed two types of MoE blocks. Table~\ref{tab:Moe} presents an analysis of the performance of different MoE block options. Specifically, we used SwinIR as the baseline and tested three configurations: 1) SwinIR with only H-MoE block; 2) SwinIR with only S-MoE block, where the expert number of S-MoE layer is set to 16; 3) SwinIR with both S-MoE and H-MoE blocks arranged alternately. We observe that both H-MoE and S-MoE significantly enhance performance when compared to the baseline. For H-MoE, the FLOPs remain constant, but the number of parameters increases. For S-MoE, both the number of parameters and FLOPs greatly increase. When H-MoE and S-MoE are alternated, the performance further improves while achieving a better balance between H-MoE and S-MoE in terms of parameter number and FLOPs. Therefore, we default to using this alternated design.

 \noindent\textbf{Single to multiple tasks.} In this experiment, we want to study the performance influence of the multi-task and multi-level support in the unified model. In Table~\ref{table:single_to_multi_task}, we use the SwinIR trained on each specific task and degradation level as the baseline. It shows that, despite the flexibility, using one unified model to support multiple restoration tasks and multiple degradation levels will both result in the performance drop for each specific evaluation setting. By integrating the MoE mechanism that uses different experts to decouple different tasks, our proposed DAMoE can greatly improve the performance.   

\section{Conclusions and Limitations}
In this paper, we present \dataset~, a large-scale, high-quality dataset that is suitable for various image restoration tasks. This effort is crucial in advancing image restoration tasks to a larger scale and unlocking the full potential of restoration models. While the number of images in \dataset~ is larger than most existing low-level datasets, it is still several orders of magnitude smaller than datasets designed for other understanding tasks, such as Laion-5B \cite{schuhmann2021laion} for multimodal pretraining. In the near future, we plan to further expand the dataset to an even larger scale. Based on \dataset, we also propose a unified restoration model DAMoE, which incorporates the mix of expert idea to handle different restoration tasks simultaneously. Our extensive experimental results demonstrate the value of our proposed dataset and unified model.

{\small
\bibliographystyle{ieee_fullname}
\bibliography{egbib}
}
\clearpage
\appendix
\section{DataSet}
\subsection{Rich Texture Details}
\label{sec:rich_texture_details}
As emphasized in our main text, the level of detail in the image's texture plays a crucial role in determining the quality of a dataset. To accurately measure this, we employ a quantitative method by calculating the high-frequency ratio of each image. This ratio indicates the proportion of high-frequency components in the image, which directly correlates with the richness of texture details. To calculate the high-frequency ratio, we begin by mapping the image to the frequency domain using Discrete Fourier Transform (DFT), as shown in Equation (\ref{equation:dft}). Next, we apply the Ideal High-pass Filter (IHPF), represented by Equation (\ref{equation:ihpf}), to extract the high-frequency components of the image. Finally, we compute the ratio of the power spectrum of the high-frequency components, as given by Equation (\ref{equation:ratio}).

\begin{align}
    \centering
   & F(u,v) = \frac{1}{HW}\sum_{x=0}^{H-1}\sum_{y=0}^{W-1}f(x,y)e^{-2j\pi(ux/H+vy/W)}\label{equation:dft}
\end{align}
where $H,W$ is the height and width of the image, $f(x,y)$ is the gray value corresponding to the pixel with coordinates (x, y).
\begin{align}
    \centering
   & D(u,v) = [(u-\frac{H}{2})^2+(v-\frac{W}{2})^2]^{\frac{1}{2}}\label{equation:distance}\\
   & H(u,v) = \begin{cases}
   0,&D(u,v)\leq D_0 \\
   1,&D(u,v)\textgreater D_0\\\end{cases}\label{equation:ihpf}
\end{align}
\begin{align}
   &ratio = \frac{\sum_{(u,v)}|H(u,v) \cdot F(u,v)|^2}{\sum_{(u,v)}|F(u,v)|^2}\label{equation:ratio}
\end{align}
where $D(u,v)$ is the distance between the frequency point $(u,v)$ and the center of the frequency domain$(\frac{H}{2},\frac{W}{2})$, $D_0$ is the cutoff frequency of the ideal high-pass filter, in this paper , we set $D_0=\frac{1}{2}\sqrt{(\frac{H}{2})^2+(\frac{W}{2})^2}$

\subsection{Semantic Coverage}
\label{sec:semantic}
\begin{figure*}
    \centering
    \includegraphics[width=0.95\textwidth]{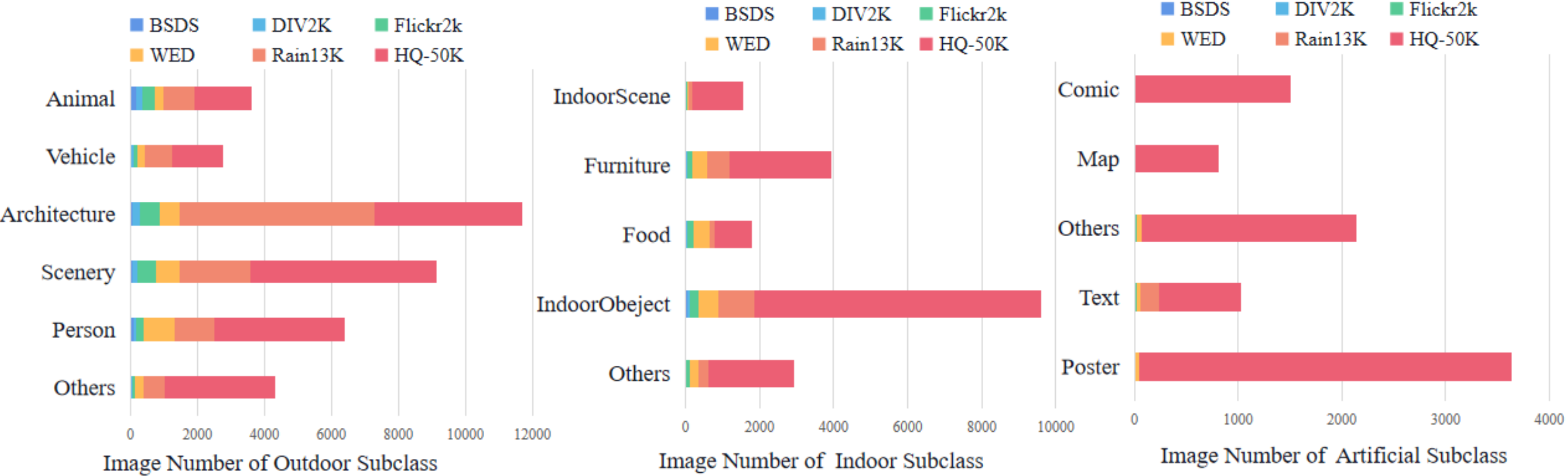}
    \caption{Semantic distribution of datasets. (a), (b), (c) show the image number of each subclass in Outdoor, Indoor, Artificial class respectively.}
    \label{fig:subclass}
\end{figure*}

In our main text, we discussed how the existing restoration datasets have inferior semantic coverage compared to our dataset, especially for the artificial images. In Figure \ref{fig:subclass}, we present a detailed distribution of the image numbers across different semantic sub-categories for both our dataset and the existing ones. It can be seen that there is few or no images for the subclass of map, comic or posters in existing restoration datasets. Besides, we also maintain a better balance of different sub-categories in our \dataset~ to avoid the performance bias. In Table \ref{tab:fine-grained}, we show the detailed performance of SwinIR trained on DIV2K and our \dataset~ for each sub-category (The summarized average performance is reported in Table 4 in our main text). It can be observed that the performance gap of the model trained on our \dataset~ and DIV2K\cite{DIV2K} is significantly larger on the artificial subclass due to the limited artificial image coverage of DIV2K\cite{DIV2K}.

Note that due to the diverse semantic categories, it is challenging to categorize each image into a well-defined category set. For images that do not belong to these subcategories, we classify them into ``others” For instance, for Outdoor, ``others" include traffic lights, road sign devices, carry-on items, \etc.; for Indoor, ``others” includes knitting textures, electronic devices \etc.; for Artificial, ``others" include game interfaces, handwriting, screen shots \etc.

\begin{figure*}[t]
\centering
    \includegraphics[width=\textwidth]{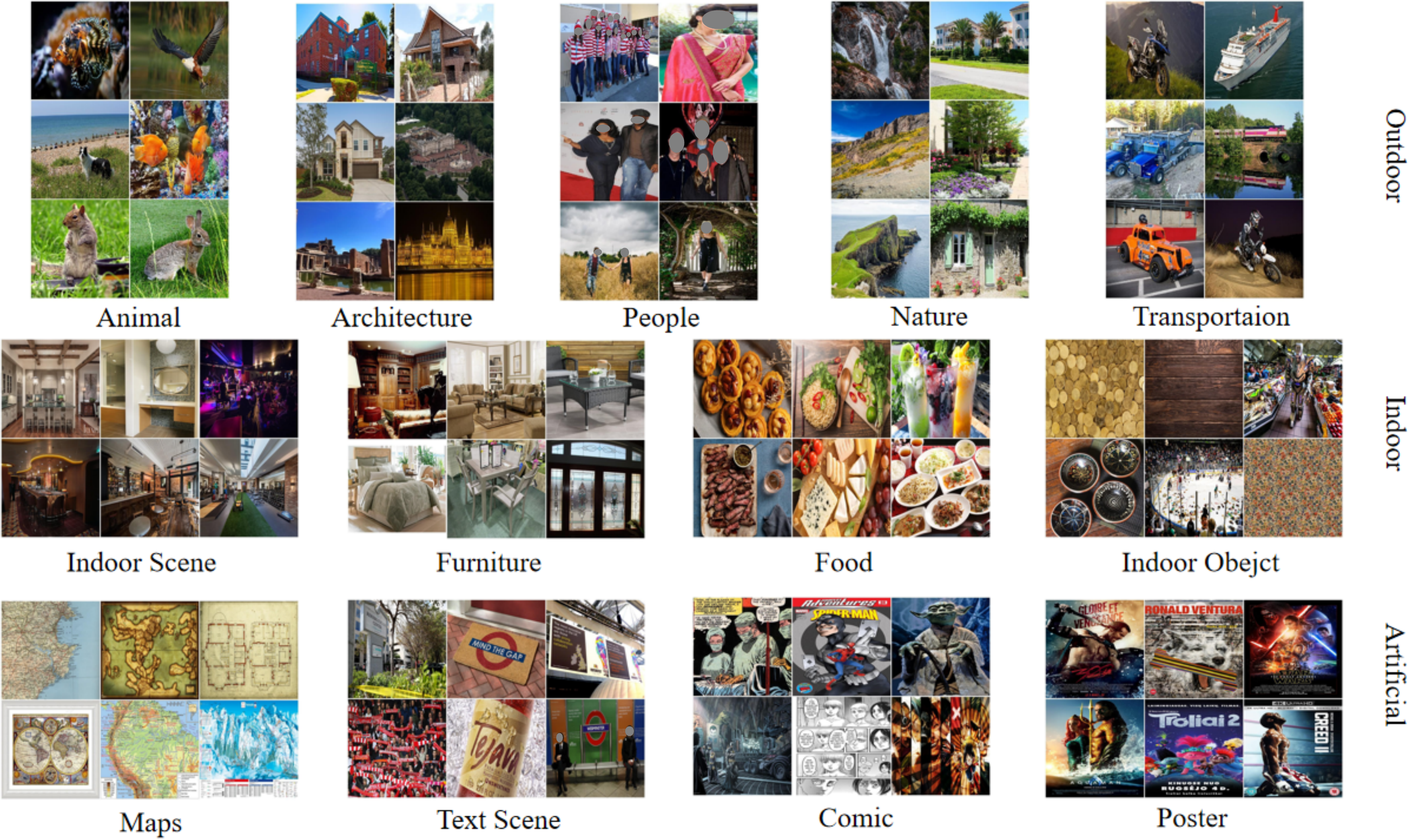}
    \caption{Example images in our fine-grained benchmark dataset. From top to down are the images of Outdoor, Indoor, and Artificial respectively. }
    \label{fig:visualize}
\end{figure*}

For the fine-grained test benchmark we proposed, we give some visualization examples in Figure \ref{fig:visualize}. Except for the text scene, which contains 50 test images, all other 12 sub-categories have 100 test images. 

\begin{table*}[]
\centering
\caption{Comparison of the SR results trained on DIV2K and our \dataset~ on our fine-grained evaluation benchmark: Outdoor(top), Indoor(middle), Artificial(bottom). }
\label{tab:fine-grained}
\setlength{\tabcolsep}{3.5pt}{
\begin{tabular}{c|c|ll|ll|ll|ll|ll|ll}
\hline
\multirow{2}{*}{Scale} &\multirow{2}{*}{TrainingSet} & \multicolumn{2}{c|}{Animals} & \multicolumn{2}{c|}{People} & \multicolumn{2}{c|}{Nature} & \multicolumn{2}{c|}{Transportation} & \multicolumn{2}{c|}{Architecture} & \multicolumn{2}{c}{Outdoor} \\\cline{3-14}
                       &                              & PSNR        & SSIM         & PSNR        & SSIM         & PSNR         & SSIM         & PSNR         & SSIM         & PSNR           & SSIM            & PSNR         & SSIM         \\\hline
\multirow{2}{*}{x2}     & DIV2K                                            & 35.30          & 0.9169          & 30.19          & 0.8907          & 28.07          & 0.8770          & 31.47          & 0.9206          & 28.84          & 0.8988          & 30.78          & 0.9008          \\
                        & HQ-50K                                           & \textbf{35.40} & \textbf{0.9194} & \textbf{30.50} & \textbf{0.8979} & \textbf{28.20} & \textbf{0.8827} & \textbf{31.67} & \textbf{0.9256} & \textbf{29.23} & \textbf{0.9078} & \textbf{31.00} & \textbf{0.9067} \\\hline
\multirow{2}{*}{x3}     & DIV2K                                            & 31.68          & 0.8337          & 27.22          & 0.7995          & 25.05          & 0.7540          & 28.14          & 0.8413          & 25.63          & 0.7939          & 27.54          & 0.8045          \\
                      & HQ-50K                                           & \textbf{31.84} & \textbf{0.8352} & \textbf{27.34} & \textbf{0.8032} & \textbf{25.10} & \textbf{0.7569} & \textbf{28.31} & \textbf{0.8447} & \textbf{25.88} & \textbf{0.8029} & \textbf{27.69} & \textbf{0.8086} \\\hline
\multirow{2}{*}{x4}    & DIV2K                                            & 29.81          & 0.7724          & 25.80          & 0.7368          & 23.67          & 0.6642          & 26.50          & 0.7809          & 24.08          & 0.7126          & 25.97          & 0.7334          \\
                      & HQ-50K                                           & \textbf{29.99} & \textbf{0.7775} & \textbf{25.98} & \textbf{0.7413} & \textbf{23.78} & \textbf{0.6680} & \textbf{26.74} & \textbf{0.7862} & \textbf{24.39} & \textbf{0.7214} & \textbf{26.18} & \textbf{0.7389}
                       \\\hline
\end{tabular}
}

\vspace{1em}
\setlength{\tabcolsep}{6.2pt}{
\begin{tabular}{c|c|ll|ll|ll|ll|ll}
\hline
\multirow{2}{*}{Scale} &\multirow{2}{*}{TrainingSet} & \multicolumn{2}{c|}{Indoor Scenes	
} & \multicolumn{2}{c|}{Furniture} & \multicolumn{2}{c|}{Food} & \multicolumn{2}{c|}{Indoor Object} & \multicolumn{2}{c}{Indoor}  \\\cline{3-12}
                       &                              & PSNR        & SSIM         & PSNR        & SSIM         & PSNR         & SSIM         & PSNR         & SSIM         & PSNR           & SSIM                  \\\hline
\multirow{2}{*}{x2}     & DIV2K                                            & 31.56          & 0.9079          & 30.63          & 0.9091          & 33.73          & 0.9207          & 28.93           & 0.8835           & 31.21          & 0.9053          \\
                    & HQ-50K                                           & \textbf{31.90} & \textbf{0.9129} & \textbf{31.12} & \textbf{0.9176} & \textbf{33.91} & \textbf{0.9272} & \textbf{29.15}  & \textbf{0.8961}  & \textbf{31.52} & \textbf{0.9134} \\\hline
\multirow{2}{*}{x3}   & DIV2K                                            & 28.43          & 0.8316          & 27.51          & 0.8289          & 30.37          & 0.8476          & 25.68           & 0.7721           & 28.00          & 0.8201          \\
                       & HQ-50K                                           & \textbf{28.63} & \textbf{0.8321} & \textbf{27.80} & \textbf{0.8311} & \textbf{30.51} & \textbf{0.8501} & \textbf{25.90}  & \textbf{0.7765}  & \textbf{28.21} & \textbf{0.8224} \\\hline
\multirow{2}{*}{x4}    & DIV2K                                            & 26.84          & 0.7757          & 25.96          & 0.7686          & 28.68          & 0.7925          & 24.12           & 0.6862           & 26.40          & 0.7558          \\
                       & HQ-50K                                           & \textbf{27.01} & \textbf{0.7811} & \textbf{26.24} & \textbf{0.7762} & \textbf{28.82} & \textbf{0.7951} & \textbf{24.36}  & \textbf{0.6934}  & \textbf{26.60} & \textbf{0.7614}     
                       \\\hline
\end{tabular}
}
\vspace{1em}
\setlength{\tabcolsep}{6.2pt}{
\begin{tabular}{c|c|ll|ll|ll|ll|ll}
\hline
\multirow{2}{*}{Scale} &\multirow{2}{*}{TrainingSet} & \multicolumn{2}{c|}{Comic} & \multicolumn{2}{c|}{Poster} & \multicolumn{2}{c|}{Map} & \multicolumn{2}{c|}{Text Scene} & \multicolumn{2}{c}{Artificial}  \\\cline{3-12}
                       &                              & PSNR        & SSIM         & PSNR        & SSIM         & PSNR         & SSIM         & PSNR         & SSIM         & PSNR           & SSIM             \\\hline
\multirow{2}{*}{x2}    & DIV2K                                            & 31.10          & 0.9418          & 32.22          & 0.8977          & 27.77          & 0.9053          & 29.96          & 0.8941          & 30.31          & 0.9120          \\
                      & HQ-50K                                           & \textbf{31.35} & \textbf{0.9600} & \textbf{32.46} & \textbf{0.9178} & \textbf{28.06} & \textbf{0.9172} & \textbf{30.01} & \textbf{0.9072} & \textbf{30.47} & \textbf{0.9256} \\\hline
\multirow{2}{*}{x3}   & DIV2K                                            & 26.72          & 0.8864          & 28.96          & 0.8302          & 24.17          & 0.8163          & 26.82          & 0.8029          & 26.65          & 0.8384          \\
                      & HQ-50K                                           & \textbf{27.86} & \textbf{0.8997} & \textbf{29.45} & \textbf{0.8355} & \textbf{25.26} & \textbf{0.8356} & \textbf{27.18} & \textbf{0.8114} & \textbf{27.47} & \textbf{0.8504} \\\hline
\multirow{2}{*}{x4}   & DIV2K                                            & 24.23          & 0.8314          & 27.25          & 0.7841          & 22.58          & 0.7439          & 25.44          & 0.7406          & 24.79          & 0.7799          \\
                      & HQ-50K                                           & \textbf{25.38} & \textbf{0.8520} & \textbf{27.79} & \textbf{0.7915} & \textbf{23.41} & \textbf{0.7697} & \textbf{25.72} & \textbf{0.7499} & \textbf{25.55} & \textbf{0.7966}                         \\\hline
\end{tabular}
}
\end{table*}

\begin{figure*}
    \centering
    \includegraphics[width=0.95\textwidth]{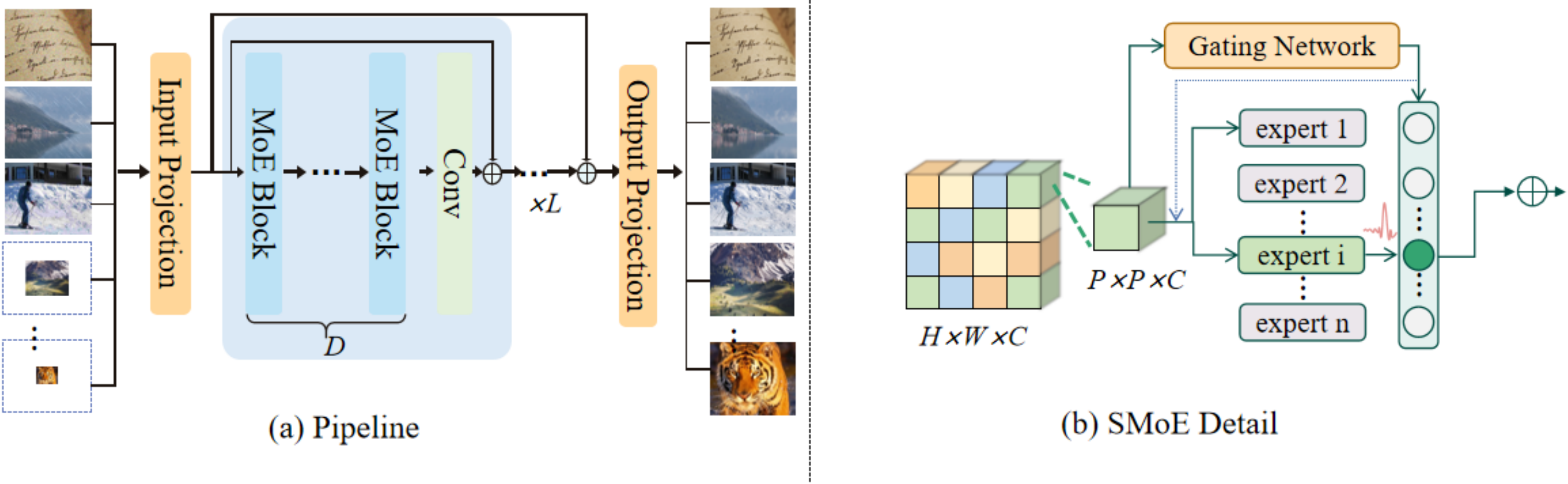}
    \caption{More details about the network. (a) shows the pipeline. (b) introduces the patch-level tokens in SMoE}
    \label{fig:sup_network}
\end{figure*}

\section{Network Structure}
\label{sec:network}
\subsection{Detailed Network Architectures}
\label{sec:Skip-connection}
In this section, we will provide a detailed explanation of our DAMoE network framework for unified restoration. A total of 11 degradations are involved in our unified training, including SR for  $\times2$, $\times3$, $\times4$, deraining, denoising for $\sigma = \left\{ 15,25,50 \right\}$, dejpeg for $q=  \left\{ 10,20,30,40 \right\}$). The overall structure is shown in Figure~\ref{fig:sup_network}, where the number of MoE-conv Blocks $L$, the number of MoE blocks $D$ , the input size, window size, channel number $C$ and attention head number are set to 6, 6, $84\times84$, 7, 180 and 6 respectively. In H-MoE (Hard-MoE), the number of experts is the same as the number of tasks, which is $4$ in our experiment, while the number of SMoE (Soft-MoE) is set to $16$ to achieve a trade-off of performance and computing cost.

\subsection{Detailed Design of HMoE and SMoE}
\label{sec:Patch-level}
During our early attempts, we found that a simple combination of Transformer and MoE layers did not perform well in low-level tasks. The possible reason is that, in the backbone SwinIR~\cite{swinir} that we select as a baseline, each token corresponds to a pixel rather than a patch, when it comes to the gating network of MoE, the input token corresponding to a single pixel is too small to be conducive to the selection of the gating. Without gating constraints, it was more challenging for the model to converge compared to the typical patch-wise tokens. To address this issue, we first simplified the MoE design to HMoE. In H-MoE, the input feature $x$ is visible only to the corresponding task-specific expert, avoiding the problem of limited receptive field of the gating network.

To take advantage of gating strategy and solve the problem, we further propose patch-level tokens in SMoE. In detail, as Figure \ref{fig:sup_network} (b) shows, given the 2D feature maps $X \in \mathbb{R}^{C \times H \times W}$, where $H$ and $W$ are the height and width of the feature, we first split $X$ into no-overlapping patches with the patch size of $P \times P$, and then obtain the $i$th flatten embedding $x_i \in \mathbb{R}^{P^2 \times C}$ from each patch $i$:
\begin{align}
    \centering
   & X = \left\{ x_1,x_2,...,x_N \right\}, N = HW/P^2\\
\end{align}
Then for each $x_i$, we feed it into SMOE:
\begin{align}
    \centering
   & w_i = \mathcal{Z}_k(\textrm{softmax}(\mathbf{W_g}\otimes x_i)),\label{eq:W}\\
   & y_i = \sum_j^n{w(x_i)_j E_j(x_i)}, \label{eq:y}\\
   & Y = \left\{y_1,y_2,...,y_N\right\}, N = HW/P^2
\end{align}
where $n$ is the number of expert, $w_i \in \mathbb{R}^{n}$ is the weight of different experts for the input $x_i$. As described in the main text, the $w(x_i)$ is obtained by the gate network in eq. (\ref{eq:W}). $\mathbf{W_g} \in \mathbb{R}^{n \times C}$ is a trainable parameter, $\otimes$ is the matrix multiplication, $\mathcal{Z}_k(\cdot)$ is the function that sets all values to zeros except the top-$k$ largest values. In this paper, we set $P=7$ and $k=1$, which means that for a SMoE Block in eq.(\ref{eq:y}) , only the activated expert $E_{j}(x_i)$ contributes to  the output $y_i=w(x_i)_jE_{j}(x_i)$. Finally, we merge the output of each input patch, and reshape the flatten embedding back to get the $Y \in \mathbb{R}^{C \times H \times W}$. 
\begin{figure*}
    \centering
    \includegraphics[width=0.95\textwidth]{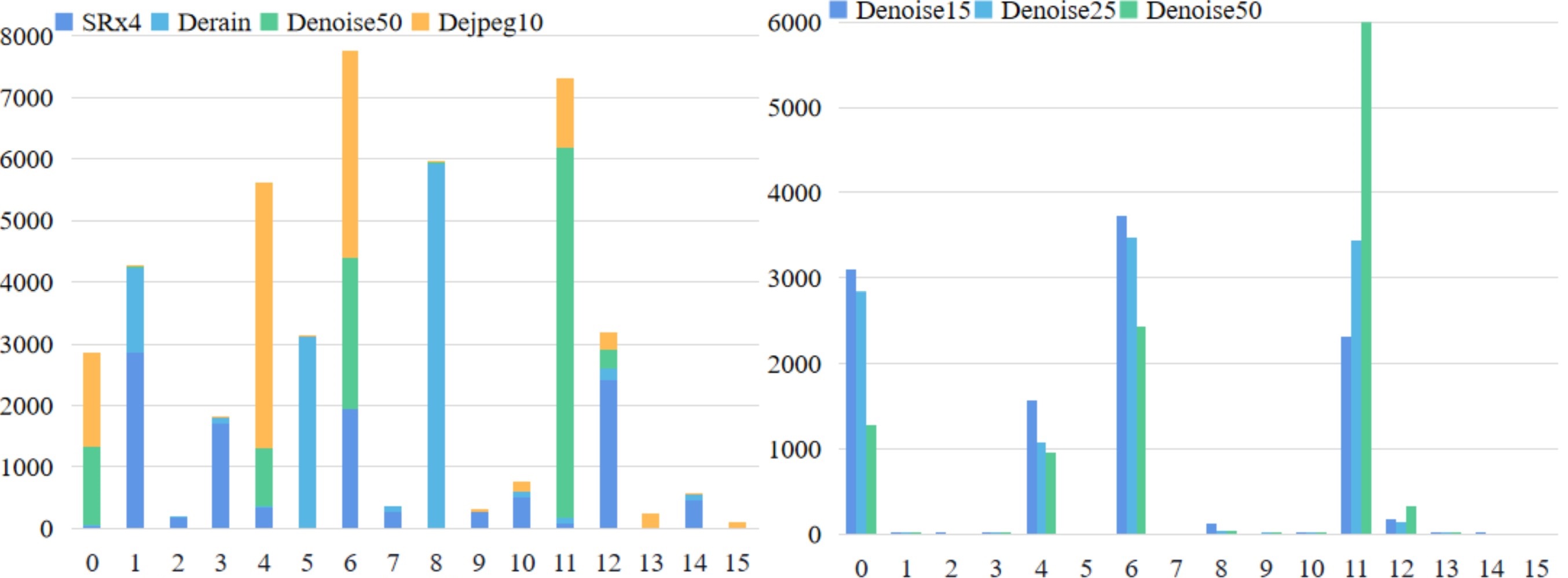}
    \caption{Expert selection pattern of different Task (left) and different degradation levels of the same task (right).}
    \label{fig:expert_router}
    \includegraphics[width=0.95\textwidth]{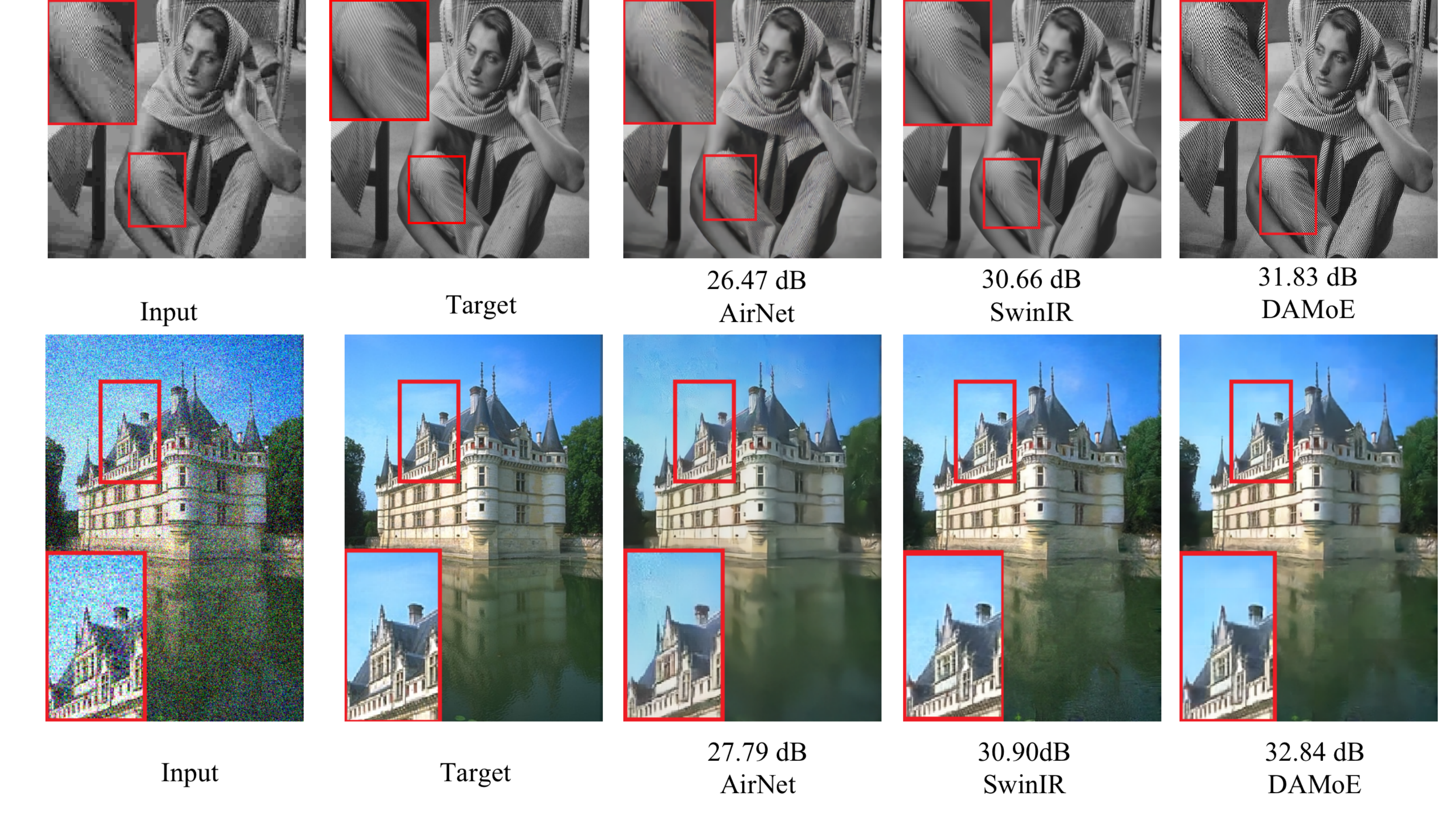}
    \caption{Visual comparisons of different methods on Dejpeg of $q=10$ on classic5 \cite{classical5} and on Denoise of $\sigma=50$ on CBSD68 \cite{CBSD68}}
    \label{fig:denoising_visual}
\end{figure*}

\section{More Results}
\subsection{More comparison Under the Same Data Scale.}\label{sec:scale_factor}
To further demonstrate the value of the proposed dataset, we compared DAMoE's performance using ImageNet-50K \cite{ImageNet}, Laion-50K \cite{schuhmann2021laion}, and our proposed \dataset~. For ImageNet-50K and Laion-50K, we randomly selected 50,000 images from each dataset. The results are presented in Table \ref{tab:50K}. As shown, our \dataset~ outperforms the other two datasets across all degradation levels, indicating its superior quality.

\begin{figure*}[!t]
    \centering
    \includegraphics[width=0.95\textwidth]{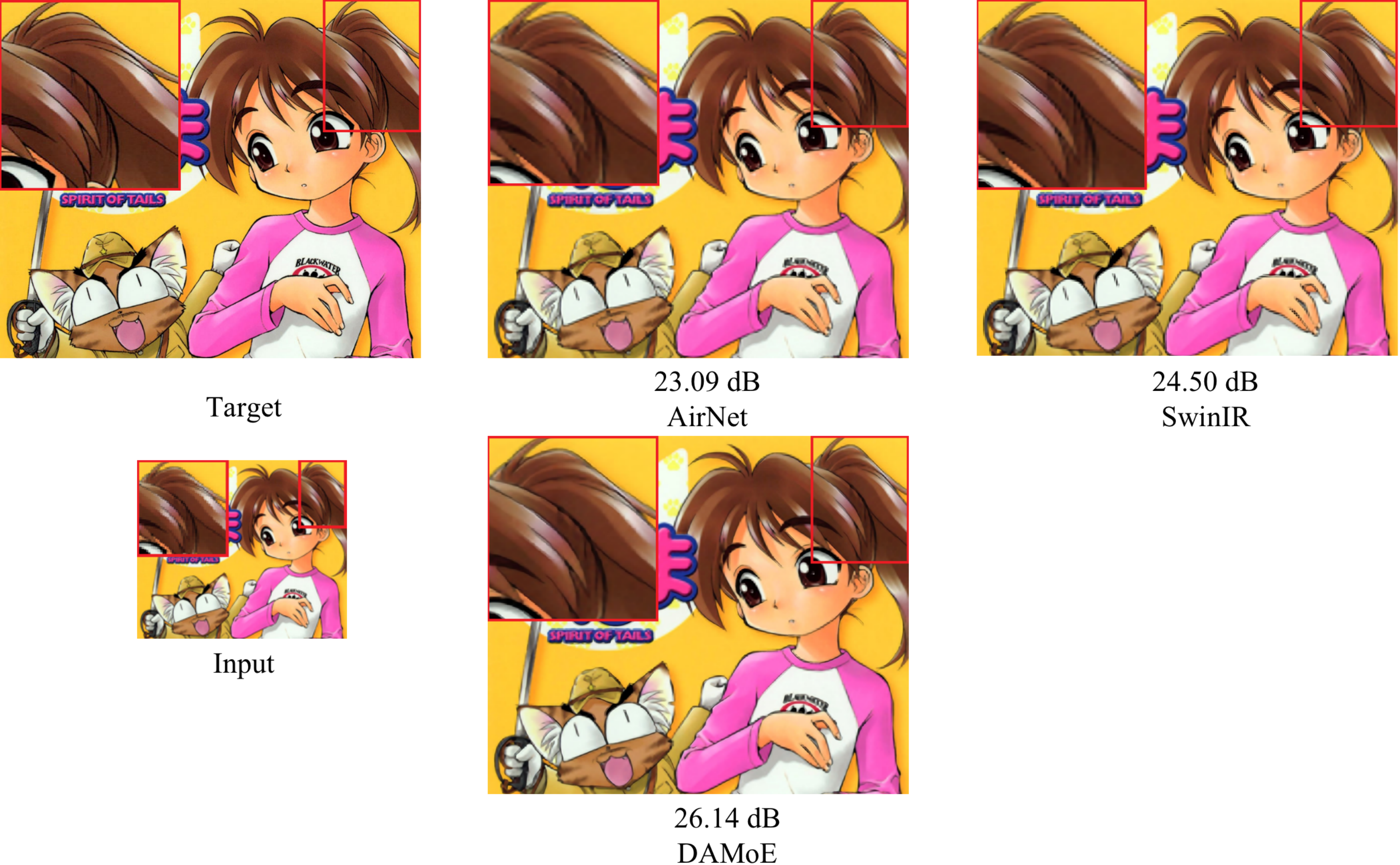}        \caption{Visual comparisons of different methods on SR of $\times4$ on Manga109 \cite{manga109}}
    \label{fig:sr_visual}
    
    \includegraphics[width=0.95\textwidth]{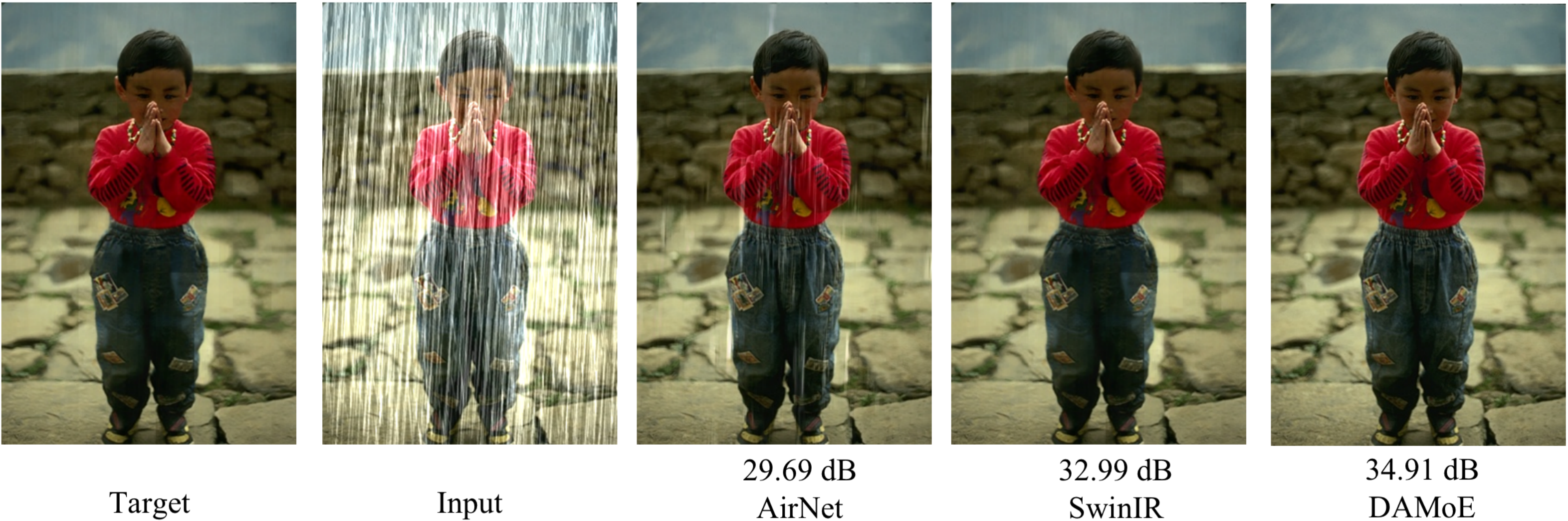}
    \caption{Visual comparisons of different methods on DeRain on Rain100 \cite{rain100}}
        \label{fig:derain_visual}
\end{figure*}

\begin{table*}[!t]
\centering
\caption{Comparisons with the multitask results in ALL-in-One DAMoE model trained on ImageNet-50K, Laion-50K and our \dataset~}
\label{tab:50K}
\setlength{\tabcolsep}{6pt}{
\begin{tabular}{l|ccc|c|ccc|cccc}
\hline
\multirow{3}{*}{Training Set} & \multicolumn{3}{c}{SR(PSNR-Y)} &  \multicolumn{1}{|c}{DeRain(PSNR-Y)} & \multicolumn{3}{|c}{Denoise(PSNR)} & \multicolumn{4}{|c}{Dejpeg(PSNR)} \\
                                          & \multicolumn{3}{c}{Manga109\cite{manga109}}  & \multicolumn{1}{|c}{Rain100L\cite{rain100} }     & \multicolumn{3}{|c}{CBSD68\cite{CBSD68}}       & \multicolumn{4}{|c}{Classic5\cite{LIVE1}}       \\
                              \cline{2-12}
                                                     & x2       & x3       & x4      & -              & 15        & 25        & 50       & 40     & 30     & 20    & 10    \\ 
                              \hline
                          Laion-50K                                  & 38.22          & 33.13          & 25.24          & 39.07          & 34.12          & 31.47          & 28.24          & 33.99          & 33.18          & 31.93          & 29.74          \\
ImageNet-50K                                  & 37.27          & 32.32          & 25.53          & 38.59          & 34.01          & 31.36          & 28.10          & 33.94          & 33.10          & 31.83          & 29.61          \\
HQ-50K                                       & \textbf{38.51} & \textbf{33.40} & \textbf{26.12} & \textbf{40.22} & \textbf{34.18} & \textbf{31.52} & \textbf{28.28} & \textbf{34.15} & \textbf{33.31} & \textbf{32.03} & \textbf{29.80} \\\hline
\end{tabular}
}
\end{table*}

\subsection{Adaptive Expert Routing of SMoE}
\label{sec:SMoE Routing}
We further analyzed the expert selection pattern of SMoE in the MoE block. To this end, we selected the SMoE in the $18th$ MoE Block, i.e., the SMoE in the middle of the network, as the analysis object, as illustrated in Figure \ref{fig:expert_router}. Our findings revealed that for various tasks, the model will make different selections of experts. Moreover, even for the same task with different degradation levels, there are differences in expert routing, albeit not as pronounced as those observed between different tasks. Our results align with the domain distance of its degraded domain.

\subsection{Visual Results}
\label{visual_results}
We provide some qualitative comparisons to show the superiority of our DAMoE as the unified restoration model. The Figure \ref{fig:denoising_visual},\ref{fig:sr_visual},\ref{fig:derain_visual} show the result of SR, deraining, dejpeg and denoising that all the model are trained on our \dataset~ respectively. For SR, our method suffers from less ringing effects and has sharper texture compared with the others. For deraining, our DAMoE can better remove dense rain streaks. For dejpeg, our results preserve more high-frequency information. And for denoising, DAMoE could restore a cleaner image. This illustrates the effectiveness of integrating the MoE idea for unifying different restoration tasks.

\end{document}